\documentclass{article}

\usepackage{microtype}
\usepackage{graphicx}
\usepackage{booktabs} %

\usepackage{hyperref}

\usepackage[accepted]{icml2025}

\usepackage{amsmath}
\usepackage{amssymb}
\usepackage{mathtools}
\usepackage{amsthm}

\usepackage[capitalize,noabbrev]{cleveref}

\theoremstyle{plain}

\theoremstyle{definition}

\theoremstyle{remark}

\usepackage[textsize=tiny]{todonotes}

\usepackage{xspace}
\usepackage{nicefrac}
\usepackage{subcaption}
\usepackage{caption}
\usepackage{arydshln}
\usepackage{placeins}

\newcommand{\methodname}{FastCAV\xspace} %
\newcommand{\oldmethodname}{SVM-CAV\xspace}

\newcommand{\maxspeedup}{$63.6\times$\xspace}
\newcommand{\avgspeedup}{$46.4\times$\xspace}
\newcommand{\concept}[1]{``#1''\xspace}

\usepackage{marginnote}

\icmltitlerunning{\methodname: Efficient Computation of Concept Activation Vectors}

\begin{document}

\twocolumn[
\icmltitle{\methodname: Efficient Computation of Concept Activation Vectors \\ for Explaining Deep Neural Networks}

\icmlsetsymbol{equal}{*}

\begin{icmlauthorlist}
\icmlauthor{Laines Schmalwasser}{dlr,fsu}
\icmlauthor{Niklas Penzel}{fsu}
\icmlauthor{Joachim Denzler}{fsu}
\icmlauthor{Julia Niebling}{dlr}
\end{icmlauthorlist}

\icmlaffiliation{dlr}{Institute of Data Science, German Aerospace Center, Jena, Germany}
\icmlaffiliation{fsu}{Computer Vision Group, Friedrich Schiller University Jena, Germany}

\icmlcorrespondingauthor{Laines Schmalwasser}{Laines.Schmalwasser@dlr.de}

\icmlkeywords{Machine Learning, ICML}

\vskip 0.3in
]

\printAffiliationsAndNotice{}  %

\begin{abstract}
    Concepts such as objects, patterns, and shapes are how humans understand the world.
    Building on this intuition, concept-based explainability methods aim to study representations learned by deep neural networks in relation to human-understandable concepts.
    Here, Concept Activation Vectors (CAVs) are an important tool and can identify whether a model learned a concept or not.
    However, the computational cost and time requirements of existing CAV computation pose a significant challenge, particularly in large-scale, high-dimensional architectures. 
    To address this limitation, we introduce \methodname, a novel approach that accelerates the extraction of CAVs by up to \maxspeedup (on average \avgspeedup). %
    We provide a theoretical foundation for our approach and give concrete assumptions under which it is equivalent to established SVM-based methods.
    Our empirical results demonstrate that CAVs calculated with \methodname maintain similar performance while being more efficient and stable.
    In downstream applications, i.e., concept-based explanation methods, we show that \methodname can act as a replacement leading to equivalent insights.
    Hence, our approach enables previously infeasible investigations of deep models, which we demonstrate by tracking the evolution of concepts during model training.\footnote[3]{Project page: \url{https://fastcav.github.io/}}
\end{abstract}
    
\section{Introduction}
\begin{figure}[t]
    \centering
    \includegraphics[width=0.95\linewidth]{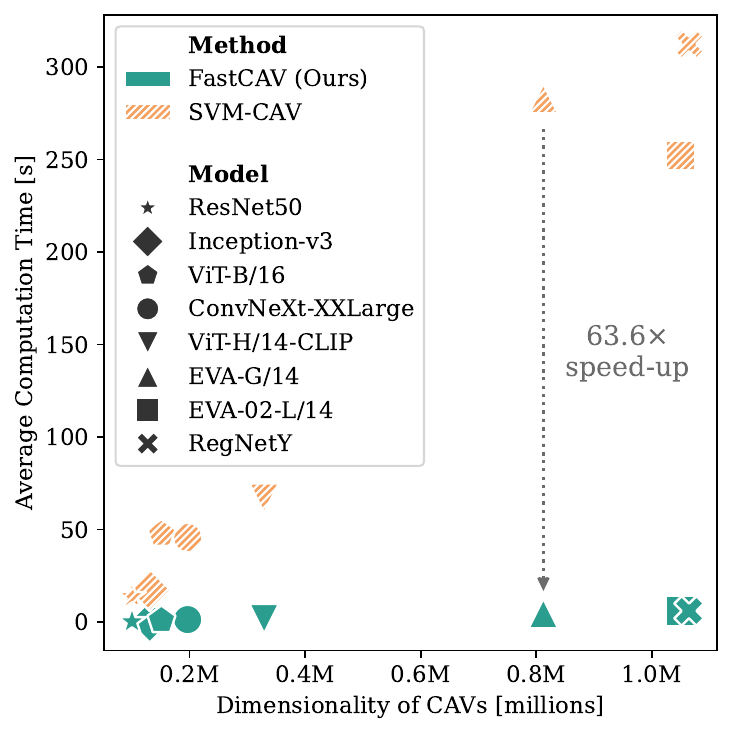}
    
    \caption{
    Comparison of computational efficiency between \methodname and established \oldmethodname for calculating Concept Activation Vectors (CAVs) across different models.
    The average time to calculate a CAV for each method is plotted against the dimensionality of the activation spaces, demonstrating the significant speedup ($p < 0.01$) achieved by \methodname. 
    }
    \label{fig:introfigure}
\end{figure}

Humans often think in terms of abstract, perceptually grounded concepts to make sense of the world \cite{barsalou1999perceptual}. 
This led to the idea that explaining models in the same conceptual terms makes their behavior more intuitive and relatable to human understanding \cite{kim2018interpretabilityfeatureattributionquantitative,ghorbani2019towards,yeh2020completeness,zhang2021invertible,pfau2021robust,nicolson2024explainingexplainabilityunderstandingconcept}.
Concept-based methods investigate the learned representations of a model with respect to human-understandable concepts and quantify the influence of these concepts on the predictions of a selected class.
For example: are the concepts ``stripes'', ``wheels'', or ``green'' relevant for the prediction of class \emph{zebra}?
To answer such questions, Concept Activation Vectors (CAVs) are introduced in \cite{kim2018interpretabilityfeatureattributionquantitative}.
CAVs are vectors in the network's activation space that represent human-understandable concepts and are the foundation for many concept-based explainability methods \cite{ghorbani2019towards, yeh2020completeness, oikarinen2024linear}.
Specifically, these vectors encode directions in a model's activation space, which correspond to changes in a specific semantic concept.
For a given class, it is possible to test whether the product of the gradient and vector is positive or negative.
If it is positive, %
the identified concept influences the class.
This is called Testing with Concept Activation Vectors (TCAV), \cite{kim2018interpretabilityfeatureattributionquantitative}.
Following the idea that concepts are encoded as linear directions in the activation space, e.g., \cite{szegedy2013intriguing,karpathy2015visualizing,goh2016decoding,alain2016understanding,bau2017network,bau2020understanding,elhage2022toy}, existing works find CAVs by optimizing linear classifiers, usually a linear support vector machine (SVM) \cite{kim2018interpretabilityfeatureattributionquantitative}.
Further, it is often necessary to calculate multiple CAVs to ensure statistical significance when quantifying the influence on a specific class \cite{kim2018interpretabilityfeatureattributionquantitative}.

Unfortunately, these conditions result in long computation times when performing CAV-based analyses. 
This problem is further exacerbated due to the increasing number of parameters and corresponding activation space dimensionalities in modern architectures.
As a concrete example, consider activations after the last convolutional layer of a ResNet50 \cite{he2016deep}, which has 100.352 dimensions.
In comparison, the number of last layer activations in recent models \cite{timm2025leaderboard}, e.g., EVA-02-L/14 \cite{fang2024eva}, reaches up to 1.049.600 dimensions.
This development and the need for multiple CAV computations result in long processing times and substantial memory consumption.
Hence, these constraints can be restrictive for many downstream applications, e.g., when analyzing many layers of a model \cite{nicolson2024explainingexplainabilityunderstandingconcept} or when interpreting numerous concepts simultaneously \cite{bau2017network}.
As a result, the practicality of using traditional CAV methods is limited when faced with modern classifiers and tasks.
A more efficient approach to compute CAVs in deeper layers without compromising the quality of the insights is needed.

In this work, we propose \methodname, an approach to speed up the extraction of CAVs within a neural network's activation space by up to \maxspeedup. 
To achieve this, we follow insights from the area of superposition \cite{goh2016decoding,elhage2022toy} that features in a model are represented as nearly orthogonal vectors in the activation space.
Hence, we find the CAV corresponding to a concept by computing the normalized mean vector of the positive concept examples.
Additionally, we show that our approach can be understood as an approximation of a Linear Discriminant Analysis (LDA) and, under certain assumptions, it is equivalent to the solution of a linear SVM \cite{shashua1999relationship}, providing an intuitive connection to previous works \cite{kim2018interpretabilityfeatureattributionquantitative}.
We empirically show that, despite the simplicity of our approach, it effectively identifies conceptual directions in high-dimensional activation spaces.
Additionally, \methodname significantly reduces computational expenses, making it more practical for real-world applications where resources and time are limited.
Finally, we apply our approach to track concepts during training and through the layers of a network, which was previously limited by our computational constraints.

The key contributions of our work are as follows:
\begin{itemize}
    \item We introduce \methodname, a novel approach for computing Concept Activation Vectors (CAVs) that significantly reduces computation time by up to \maxspeedup, making it more efficient and scalable.
    \item We provide a theoretical foundation for FastCAV and specify concrete assumptions under which it is equivalent to other linear methods.
    \item We demonstrate empirically that FastCAV produces similar CAVs to the established SVM-based approach, leading to comparable insights in downstream methods, e.g.,\cite{kim2018interpretabilityfeatureattributionquantitative,ghorbani2019towards}. 
    \item Using \methodname, we investigate the evolution of concepts during training and across different layers of a neural network.
\end{itemize}

\section{Related Work}\label{{sec:related_work}}
\paragraph{Concept Activation Vectors (CAVs)}

Alain et al. \cite{alain2016understanding} introduce the idea that features of an intermediate layer are separable using a linear classifier.
Based on this, CAVs \cite{kim2018interpretabilityfeatureattributionquantitative} are introduced, which connect abstract model representations and human-understandable concepts.
In particular, CAVs represent human-understandable concepts as directions in the activation space of a model's layer.
To compute CAVs, a linear classifier is trained to distinguish images that share a visual concept from random images.
The CAV corresponding to the concept is proportional to the normal vector of the linear decision boundary.
The assumption is that if the concept images cannot be separated in a model's activation space, the model did not adequately learn the concept.
Testing with Concept Activation Vectors (TCAV) \cite{kim2018interpretabilityfeatureattributionquantitative} uses CAVs to quantify the sensitivity of a model's predictions for a specific class to the presence of a human-understandable concept.
For example, TCAV is able to investigate questions such as whether the color ``blue'' is relevant for the model when predicting the class \emph{whale}.
Further, TCAV and, more specifically, CAVs are the foundation of various concept-based explainability methods, \cite{ghorbani2019towards,schrouff2021best,zhang2021invertible,singla2021using,gupta2024concept,ghosh2023distilling}.

Additionally, various methods have been proposed to enhance the interpretability of the discovered concepts.
In \cite{zhang2021invertible}, the authors address directional divergence, and in \cite{pfau2021robust}, the authors improve robustness against data variations.
Language-guided approaches like \cite{moayeri2023text2concept,huang2024lg}
remove the dependency on concept images. 
Instead, they are using vision-text embeddings, e.g., \cite{radford2021learning}, and map them to a selected activation space. 
Hence, they are able to extract CAVs belonging to concepts described by words.
In medical imaging, techniques have been developed for bidirectional explanations using regression concept vectors \cite{graziani2018regression}.
These approaches improved the usability of CAVs, but they remain computationally demanding due to classifier training for each CAV.
Further, a recent approach \cite{pahde2025navigating} reduces the influence of unrelated distractor patterns, resulting in a similar interpretation of the activation space.

\paragraph{Linearity in Neural Networks.}
Many works study how semantic features or concepts are encoded in neural networks as linear directions in the activation space, e.g., \cite{szegedy2013intriguing,karpathy2015visualizing,goh2016decoding,alain2016understanding,bau2017network,bau2020understanding,elhage2022toy,bricken2023decomposing,templeton2024scaling, oikarinen2024linear}.
Some identify interpretable neurons \cite{karpathy2015visualizing,bau2017network,bau2020understanding,oikarinen2022clip,oikarinen2024linear}.
In contrast, other studies suggest that neurons can also be polysemantic \cite{olah2017feature,olah2020zoom}, meaning that a neuron responds to more than one feature or concept.
In \cite{elhage2022toy}, the authors find that a network learns to encode features as a combination of neurons, i.e., a direction in feature space that does not align with the axes.
They argue that this happens if there are more features than neurons because features cannot be equally distributed across the neurons in that case.

Based on this idea, different works explore the phenomenon that a neural network can encode more features than dimensions in a hidden representation, \cite{olah2020zoom,elhage2022toy,bricken2023decomposing,templeton2024scaling}.
The strategy %
to achieve this is called \emph{superposition} \cite{olah2020zoom}.
Elhage et al. \cite{elhage2022toy} show on a synthetic dataset that a neural network in \emph{superposition} encodes the features as almost-orthogonal directions to each other.
Recent work explores superposition further for large language models \cite{bricken2023decomposing,templeton2024scaling}.

In this work, we follow the idea that features are encoded as almost orthogonal linear directions in activation space to design \methodname.

\section{Methodology}
\label{sec:method}

\subsection{Preliminaries}

Concept-based explanation approaches aim to explain trained neural networks $f$ on the level of human-understandable concepts.
Specifically, they investigate the activation space after a layer $l$.
Hence, we split $f$ into the concatenation of two mappings: $g_l$, which comprises the first $l$ layers, and $h_l$, which consists of the remaining layers.
For an input $x$, we write $f(x)=h_l(g_l(x))$.

To interpret $g_l(\cdot)$ with respect to a semantic concept $c$, we follow \cite{kim2018interpretabilityfeatureattributionquantitative} and describe a concept by a set of user-defined images $D_c$.
This visual definition allows for a specification independent of the particular training task and data.
The model $f$ has learned concept $c$ after layer $l$ if it is possible to separate images of $D_c$ from random samples of a set $D_r$ in the respective activation space. %
Specifically, in \cite{kim2018interpretabilityfeatureattributionquantitative}, the authors learn linear SVM classifiers to identify linear combinations of activations that correspond to a concept $c$.
They define a Concept Activation Vector (CAV) $v_c^l$ as the normal vector of such a learned linear decision boundary.
For binary labels $\text{sgn}(y)$ describing the membership of $x$ in $D_c$ or $D_r$, the linear classifier is %
\begin{equation} \label{eq:lin}
    y = v_c^l \cdot g_l(x) + b,
\end{equation}
where $b$ is the intersect, and $\cdot$ is the dot-product.

Intuitively, $v_c^l$ can be understood as the direction in the activation space of $g_l$ that corresponds to a concept $c$.
If the resulting classifier defined in \Cref{eq:lin} achieves a low separation performance, then $f$ has not sufficiently learned concept $c$ \cite{kim2018interpretabilityfeatureattributionquantitative}.
In the following section, we detail our alternative approach to calculating CAVs.

\subsection{\methodname}
\label{sec:fastcav}

Following insights from previous works on orthogonality in feature space, e.g., \cite{olah2020zoom,elhage2022toy}, \methodname aims to identify the direction directly as the difference between random images and a concept set.
Specifically, to find a CAV $v_c^l$ for a set of concept images $D_c$, we first calculate the mean of $g_l(x)$ for all $x \in D_c\cup D_r$, i.e., we use a maximum likelihood estimator,  \cite{bishop2006pattern}, %
\begin{equation}\label{eq:mlest}
    \hat{\mu}_{D_c\cup D_r} = \frac{1}{|D_c|+|D_r|} \sum_{x \in D_c\cup D_r} g_l(x).
\end{equation}

We now use this global mean to normalize all activation vectors $g_l(x)$ for samples in $D_c$.
Then, we follow the intuition detailed above that $v_c^l$ encodes the direction in activation space that separates samples contained in $D_c$ from samples in $D_r$.
Specifically, we calculate
\begin{equation}
    \label{eq:fastcav}
    v_c^l \propto \frac{1}{|D_c|}\sum_{x\in D_c}(g_l(x) - \hat{\mu}_{D_c\cup D_r}), \\
\end{equation}
which we normalize to unit length, following the SVM approach \cite{kim2018interpretabilityfeatureattributionquantitative}.

\begin{figure}[t]
    \centering
    \includegraphics[width=0.9\linewidth]{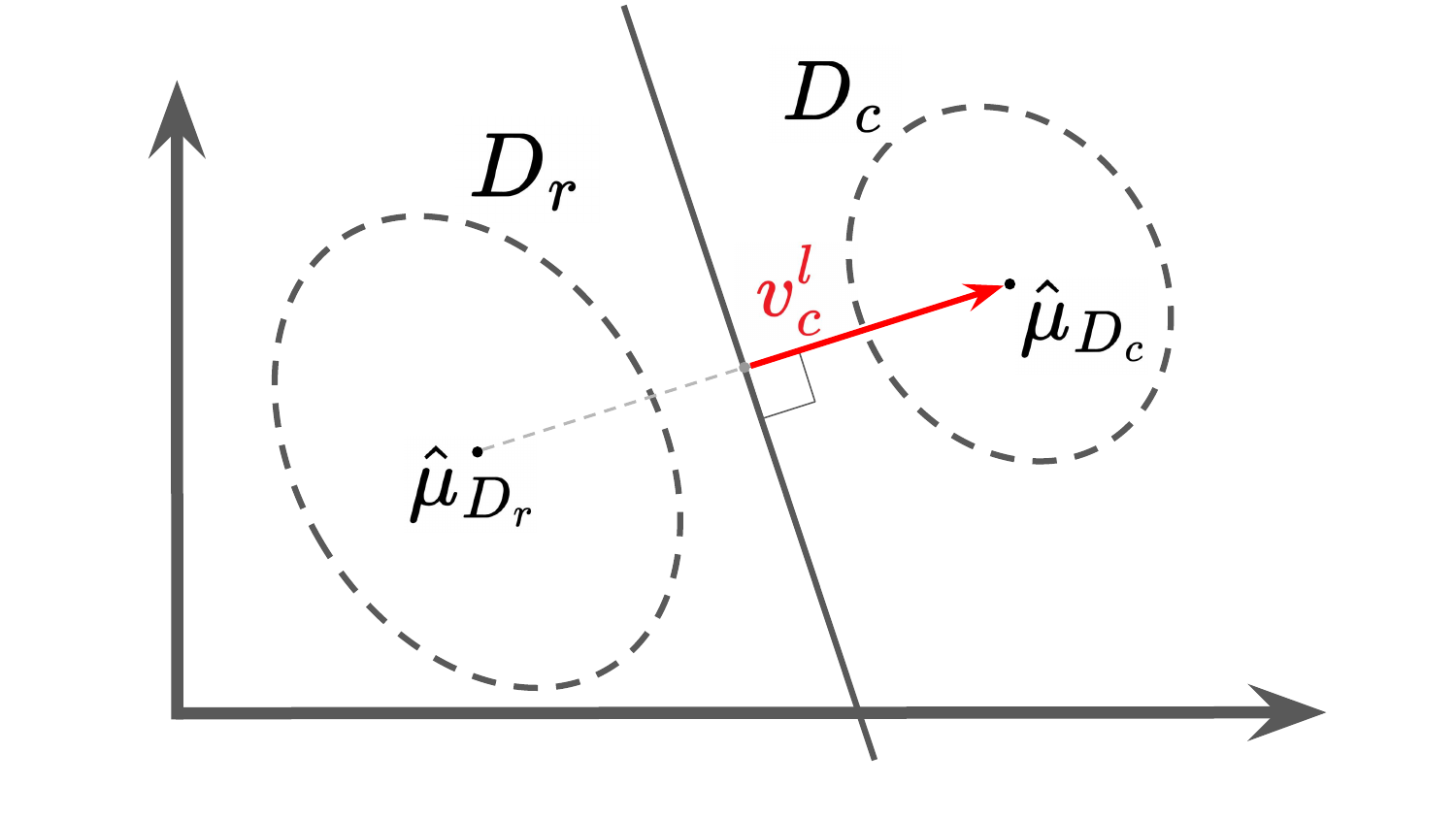}
    \caption{
    Schematic illustration of \methodname in two dimensions. 
    Specifically, the learned concept activation vector \textcolor{red}{$v_c^l$} points from the global mean $\hat{\mu}_{D_c\cup D_r}$ towards $\hat{\mu}_{D_c}$.
    }
    \label{fig:intuition}
\end{figure}

Intuitively, we first center all activations and then find the direction pointing towards the concept examples.
Hence, in \methodname, $v_c^l$ is the vector which points from $\hat{\mu}_{D_c\cup D_r}$ towards the mean of $g_l(x)$ for all $x\in D_c$, see \Cref{fig:intuition}.

With \Cref{eq:lin} and $y=0$, the intersect $b$ is given by
\begin{equation}\label{eq:class}
    b = -v_c^l \cdot \hat{\mu}_{D_c\cup D_r},
\end{equation}
fully specifying the linear decision boundary of \methodname.%

\subsection{Connection to Other Linear Classifiers}
\label{subsec:linclass}

We study \methodname on a theoretical level to investigate the connection to the established linear SVM approach for finding CAVs, e.g.,  \cite{kim2018interpretabilityfeatureattributionquantitative}.
In detail, we show that under certain assumptions, both approaches identify the same direction in latent space.
While these assumptions are strong, they provide an intuition for the strong performance of our approach in our empirical evaluations.

In \Cref{sec:fastcav}, we describe \methodname in terms of concrete observations $x \in D_c\cup D_r$.
These observations are generated by some process and follow the latent distribution of inputs.
Hence, we can study this process in terms of random variables following this unknown distribution.
Further, $g_l$ of the trained model $f$ leads to a mixture distribution for random images and images conditioned on the concept $c$.
In other words, we cast the separation of random and concept samples as a statistical learning problem, \cite{vapnik1998statistical}.

Specifically, we consider the maximum likelihood estimate of the global mean in \Cref{eq:mlest}.
This estimator is generally unbiased under \emph{i.i.d.} samples following a Gaussian distribution \cite{bishop2006pattern}.
Further, in \Cref{apx:expectation}, we show that if both the random samples and concept samples follow a multivariate Gaussian distribution and are equally mixed, then the expectation of the estimator is
\begin{equation}
\label{eq:expectation}
    \mathbb{E}[\hat{\mu}_{D_c\cup D_r}] = \frac{\mu_c + \mu_r}{2},
\end{equation}
where $\mu_c$ and $\mu_r$ are the respective means of the Gaussians.

Then \methodname finds $v_c^l$ as the unit vector proportional to the vector pointing from the global mean to the mean of the concept examples $\mu_c$.
In \Cref{eq:fastcav}, we again apply the respective maximum likelihood estimator $\hat{\mu}_{D_c}$, meaning $v_c^l \propto \hat{\mu}_{D_c} - \hat{\mu}_{D_c\cup D_r}$.
Under the assumptions specified above and again drawing an expectation over the sample of random inputs and concept examples, we get
\begin{equation}\label{eq:cav-exp}
    \mathbb{E}[v_c^l] \propto \mu_c - \frac{\mu_c + \mu_r}{2} = \frac{\mu_c-\mu_r}{2}.
\end{equation}

This solution is a simplified form of linear discriminant analysis (LDA), more specifically, Fisher discriminant analysis \cite{bishop2006pattern} under precise assumptions.
Using a related maximum likelihood estimator for the total within-class covariance 
\begin{align}
\begin{split}
    \hat{\Sigma}_{D_c\cup D_r} &= \sum_{x \in D_c} (g_l(x)-\hat{\mu}_c)(g_l(x)-\hat{\mu}_c)^\top\\
    &\hspace{0.8cm} +  \sum_{x \in D_r} (g_l(x)-\hat{\mu}_r)(g_l(x)-\hat{\mu}_c),
\end{split}
\end{align}
the Fisher discriminant solution for the normal vector of the linear decision boundary between the two classes is proportional to $\hat{\Sigma}_{D_c\cup D_r}^{-1} (\hat{\mu}_c - \hat{\mu}_r)$.
Under the explicit additional assumption of isotropic within-class covariances, i.e., $\Sigma^{-1}$ being proportional to the unit matrix, the solution for the normal vector is equivalent to \methodname.
To validate this approach, we empirically compare against the full Fisher LDA in \Cref{apx:add-comp} and find faster computation times and higher accuracies for various model architectures when using \methodname.

To summarize, under the assumptions of isotropic within-class covariances and Gaussian distributions for both random and concept examples, \methodname is equivalent to the solution of Fisher discriminant analysis on corresponding activations.
Further, in \cite{shashua1999relationship}, the authors investigate the relationship between a linear Fisher discriminant and a linear SVM, which is the established classifier utilized in \cite{kim2018interpretabilityfeatureattributionquantitative} to compute CAVs.
Specifically, they find that the solution of Fisher discriminant analysis on the set of support vectors is equivalent to the solution of the linear SVM.
Hence, a support vector machine can be understood as a sparsification of the linear Fisher discriminant.

Finally, consider that the aim is to explain the learned representation of a model after layer $l$.
Hence, these activation vectors $g_l(x)$ are potentially of a very high dimensionality $d$. %
Further, let $n = |D_c \cup D_r|$ be the number of samples and assume $d >> n$. %
Then the set of support vectors likely contains many of the samples in $D_c \cup D_r$.
Related works study this in more depth, e.g., \cite{muthukumar2021classification,hsu2021proliferation}, and find that the fraction of support vectors increases with increasing dimensionality.
We empirically confirm this and find high ratios of support vectors when studying the activations of modern architectures with linear SVM classifiers (see \Cref{tab:concept_support_vectors} in \Cref{apx:svs}).

To conclude, under the assumptions detailed above and in high-dimensional activation spaces, the normal vectors of the hyperplanes found by \methodname and a linear SVM used in \cite{kim2018interpretabilityfeatureattributionquantitative} are identical.
Further, we find that in practice, our approach leads to similar solutions while being much faster to calculate, which we demonstrate in our experimental investigations.

\subsection{Complexity Analysis}\label{subsec:complexity}

The overall time complexity for training an SVM is $\mathcal{O}(\max (n,d)\min (n,d)^2)$, where $n$ is the number of samples and $d$ is the dimensionality of the input space.
This holds for both the primal and the dual optimization problem, \cite{chapelle2007training}.
However, the complexity and speed also depend on the selected kernel, which is linear in our case  \cite{kim2018interpretabilityfeatureattributionquantitative}. 
Hence, to benchmark the speed, we utilize various common implementations of linear SVMs.
First, to solve the dual problem, we use the Sequential Minimal Optimization (SMO) algorithm \cite{platt1998fast}.
Specifically, we use the implementation in \cite{pedregosa2011scikit}, which uses the SMO implementation from \cite{chang2011libsvm}.
Second, regarding the primal problem, we use the implementation in \cite{fan2008liblinear}, which is also included in \cite{pedregosa2011scikit}.
Lastly, using Hinge loss, it is possible to approximate a linear SVM classifier using stochastic gradient descent (SGD), e.g., \cite{pedregosa2011scikit}.
For this last implementation specifically, the complexity boils down to calculating a gradient with respect to the linear decision boundary parameters ($\mathcal{O}(d)$) for each of the $n$ training samples for $T$ iterations.
Hence, the SGD variant in \cite{pedregosa2011scikit} has a complexity of $\mathcal{O}(Tnd)$.
We benchmark these three implementations and provide the results in \Cref{apx:svm-speed}.
In our use case, we find that SGD-based optimization performs best.

In comparison, \methodname calculates a sum of $n$ vectors with dimension $d$ and performs $d$ multiplications.
Hence, training complexity is $\mathcal{O}(nd)$ with very small constants.
Once the linear decision boundary is found, the inference for both \oldmethodname and \methodname is equivalent to the dot product between two $d$-dimensional vectors, resulting in $\mathcal{O}(d)$.

\section{Experiments}
\label{sec:experiments}

With \methodname, we propose an alternative to calculate CAVs, which is more computationally efficient.
To demonstrate its effectiveness, we compare our approach against the established SVM-based approach (``\oldmethodname'') \cite{kim2018interpretabilityfeatureattributionquantitative} in terms of speed and CAV quality.
We repeat this comparison for more specialized medical concepts to highlight the broad applicability.
Further, a comparison to other CAV calculation methods is provided in \Cref{apx:add-comp}, and a discussion of failure cases is provided in \Cref{apx:differences}.
Next, we show that \methodname can work as a replacement in downstream concept-based explanation methods and generate similar insights.
In particular, we compare \methodname against \oldmethodname and investigate Testing with Concepts Activation Vectors (TCAV) \cite{kim2018interpretabilityfeatureattributionquantitative}, and Automatic Concept-based Explanations (ACE) \cite{ghorbani2019towards}.
Finally, we utilize the increased efficiency of \methodname to track CAVs throughout the training and layers of a ResNet50 \cite{he2016deep} to understand how concepts evolve.
We provide the complete technical setup and comprehensive details for all our experiments in \Cref{apx:setup}.

\subsection{\methodname versus SVM-based CAV computation}\label{subsec:comparison}
To empirically compare \methodname and \oldmethodname, we evaluate a broad spectrum of architectures trained on ImageNet \cite{russakovsky2015imagenet}.
Additionally, we investigate run times for the CAV computation in a realistic setting and select current top-performing model architectures from the \texttt{timm}-leaderboard \cite{rw2019timm}. 
We used the models reported in the first column of \Cref{tab:compareFastCavwithSVMCAV}, i.e., \cite{szegedy2016rethinking,he2016deep,liu2022convnet,singh2022revisiting,dosovitskiy2020vit,fang2023eva,fang2024eva}. %

In our setup, we sample 60 concept images per set $D_c$ from the Broden dataset \cite{bau2017network}.
Similarly, we sample 60 random images from ImageNet \cite{russakovsky2015imagenet} for each set $D_r$. %
All our results are reported as averages over 30 resampled $D_r$ %
across all concepts of the Broden dataset, \cite{bau2017network}, and over the respective network layers. %
We split our investigation into four dimensions: computational time, accuracy, inter-method similarity, and intra-method robustness.

\begin{table*}%
    \small
    \centering
    \caption{Comparing our approach \methodname with \oldmethodname. 
    \textbf{Bold values} indicate better results.
    ``N/A'' indicates that no results were produced due to the overall computational time for the respective model exceeding four days.
    Additional results for other CAV calculation methods can be found in \Cref{apx:add-comp}.
    }
    \label{tab:compareFastCavwithSVMCAV}
    \begin{tabular}{lcrlrlrl:c}
        \toprule
          & & \multicolumn{2}{c}{ Comp. Time [$s$] $\downarrow$}  &\multicolumn{2}{c}{Accuracy $\uparrow$} & \multicolumn{3}{c}{Similarity $\uparrow$}\\
          \cmidrule(lr){3-4} \cmidrule(lr){5-6} \cmidrule(lr){7-9}
          Model & $\varnothing$ Dim. &  \methodname & \oldmethodname & \methodname  & \oldmethodname &  \methodname  & \oldmethodname & Inter-Method \\
        \midrule        
        Inception-v3 & 206,169 & \textbf{0.4{\scriptsize $\pm$0.14}} & 44.7{\scriptsize $\pm$29.3} & \textbf{0.95{\scriptsize $\pm$0.06}} & 0.93{\scriptsize $\pm$0.09} & \textbf{0.795{\scriptsize $\pm$0.013}} & 0.338{\scriptsize $\pm$0.062} & 0.898{\scriptsize $\pm$0.053}  \\
        ResNet50 & 341,197& \textbf{1.1{\scriptsize $\pm$1.51}} & 135.4{\scriptsize $\pm$112.9}  & \textbf{0.89{\scriptsize $\pm$0.15}} & 0.87{\scriptsize $\pm$0.15} & \textbf{0.816{\scriptsize $\pm$0.029}} & 0.400{\scriptsize $\pm$0.062} & 0.837{\scriptsize $\pm$0.146} \\
        ConvNeXt-XXL & 753,830 & \textbf{5.5{\scriptsize $\pm$3.88}} & 269.6{\scriptsize $\pm$167.0} & 0.90{\scriptsize $\pm$0.08} & \textbf{0.93{\scriptsize $\pm$0.07}} & \textbf{0.831{\scriptsize $\pm$0.042}} & 0.449{\scriptsize $\pm$0.075} & 0.807{\scriptsize $\pm$0.136} \\
        RegNetY & 1,869,479 & 8.2{\scriptsize $\pm$1.04} & N/A & 0.90{\scriptsize $\pm$0.05} & N/A & 0.734{\scriptsize $\pm$0.009} & N/A & N/A \vspace{0.15cm} \\
        
        ViT-B/16 & 140,560 & \textbf{1.1{\scriptsize $\pm$0.03}} & 50.5{\scriptsize $\pm$9.2} & \textbf{0.82{\scriptsize $\pm$0.13}} & 0.81{\scriptsize $\pm$0.14} & \textbf{0.787{\scriptsize $\pm$0.024}} & 0.408{\scriptsize $\pm$0.046} & 0.818{\scriptsize $\pm$0.064} \\
        ViT-H/14-CLIP & 281,925 & \textbf{1.9{\scriptsize $\pm$0.08}} & 69.0{\scriptsize $\pm$10.9} & \textbf{0.87{\scriptsize $\pm$0.15}} & 0.86{\scriptsize $\pm$0.15} & \textbf{0.603{\scriptsize $\pm$0.095}} & 0.249{\scriptsize $\pm$0.090} & 0.858{\scriptsize $\pm$0.083} \\
        EVA-G/14 & 696,298 & \textbf{4.7{\scriptsize $\pm$0.14}} & 248.9{\scriptsize $\pm$60.3} & \textbf{0.88{\scriptsize $\pm$0.14}} & 0.86{\scriptsize $\pm$0.14} & \textbf{0.650{\scriptsize $\pm$0.082}} & 0.250{\scriptsize $\pm$0.080} & 0.888{\scriptsize $\pm$0.060} \\
        EVA-02-L/14 & 899,653 & \textbf{6.1{\scriptsize $\pm$0.25}} & 301.6{\scriptsize $\pm$114.7} & \textbf{0.89{\scriptsize $\pm$0.14}} & 0.88{\scriptsize $\pm$0.16} & \textbf{0.675{\scriptsize $\pm$0.077}} & 0.219{\scriptsize $\pm$0.076} & 0.836{\scriptsize $\pm$0.095} \\

        \bottomrule
    \end{tabular}
\end{table*}

\paragraph{Computational time}
We analyze the theoretical time complexity of \methodname and \oldmethodname in \Cref{subsec:complexity}.
As a complementary analysis, we now compare the run time of both methods in practice. 
In \Cref{tab:compareFastCavwithSVMCAV}, columns ``Comput. time [$s$]'' report the respective average times to compute one CAV per model. %
In \Cref{fig:introfigure}, we visualize the timings for the pre-final network layers before classification.
A similar visualization for the average over all layers is included in \Cref{apx:exp1}.

Both methods require more time to calculate CAVs for models with higher activation space dimensions.
However, this observation is expected and follows from the complexity of both methods, which scale linearly in the number of dimensions $d$. 
Nevertheless, under practical conditions, \methodname is up to \maxspeedup (on average \avgspeedup) faster than \oldmethodname.

\paragraph{Accuracies of CAVs to distinguish between $D_c$ and $D_r$}\label{subsec:accuracyofCAV}
In addition to the increased computation speed of \methodname, we need to evaluate the quality of the discovered CAVs. 
Hence, we compare the predictive performance, i.e., accuracy, of the linear classifier posed by \Cref{eq:class} using CAVs obtained from %
our approach and %
\oldmethodname. 

The results are presented in \Cref{tab:compareFastCavwithSVMCAV} (``Accuracy''). 
Notably, \methodname results in slightly higher accuracies, with one exception (ConvNeXt-XXL).
Nevertheless, both methods yield similar results.
We conclude that \methodname does not compromise the predictive quality of discovered CAVs.

\paragraph{Inter-method similarity}
To investigate the agreement between the generated CAVs of both methods, we directly compare the computed CAVs. 
For that, we report in \Cref{tab:compareFastCavwithSVMCAV} (``Inter-Method'') the average cosine similarity between CAVs calculated by \methodname and by \oldmethodname. 

Across all models, we observe high similarities (values between 0.8 and 0.9).
This result indicates that both \methodname and \oldmethodname find similar directions in diverse activation spaces for a variety of concepts.

\paragraph{Intra-method robustness}\label{subsec:robustness}
Lastly, we evaluate the variation of the discovered CAVs across different $D_r$ for both methods. 
In \Cref{tab:compareFastCavwithSVMCAV}, the first two columns of ``Similarity'' report the average pairwise cosine similarity of the set of 30 CAVs aggregated across all concepts and network layers. 
 
Overall, we find \methodname computes CAVs of a higher pairwise similarity. 
This increased robustness of \methodname is due to the CAVs only differing because of the estimations of the global mean $\hat{\mu}_{D_c\cup D_r}$ while $D_c$ stays fixed.
In contrast, \oldmethodname completely relearns linear SVMs using a non-deterministic SGD implementation (see \Cref{subsec:complexity}).

\subsection{Medical Task}
While our main experiments focus on a general-purpose model and task, we show that \methodname can also be applied in more specialized scenarios.
Following prior work on medical concept extraction \cite{singla2021using}, we fine-tune a DenseNet-121 \cite{huang2017densely} on a curated subset of the MIMIC-CXR dataset \cite{johnson2019mimic-cxr}. 
Next, we use the CheXpert labeler \cite{irvin2019chexpert} to extract concept mentions from the included free-text radiology reports included with MIMIC-CXR \cite{johnson2019mimic-cxr}.
Using the resulting concepts, we derive CAVs for each of the model’s four dense blocks applying \methodname, SVM-based computation \cite{ghosh2023distilling}, and sparse logistic regression \cite{singla2021using}.
In \Cref{tab:compareFastCAVMedicalImagesSummary}, we report the average computation time, CAV accuracy, and intra-method similarity (see \Cref{subsec:comparison}).
\methodname achieves performance similar to that of the established approaches in this domain. 
However, note the strongly reduced computation time and improved similarity.
We include an evaluation of individual concepts and the technical details of our setup in \Cref{apx:medical}.

\begin{table}[!tb]
    \centering
    \caption{Comparing \methodname with SVM \cite{ghosh2023distilling} and sparse logistic regression (SL-R) \cite{singla2021using} for specialized medical concepts in a chest x-ray image classification task.
    \textbf{Bold values} indicate better results.}
    \label{tab:compareFastCAVMedicalImagesSummary}
    \begin{tabular}{lcccc}
        \toprule
        Method & C. Time [s] $\downarrow$ & Accuracy $\uparrow$ & Similarity $\uparrow$ \\
        \midrule
        FastCAV & \textbf{0.006{\scriptsize$\pm$0.001}} & 0.72{\scriptsize$\pm$0.11} & \textbf{0.79{\scriptsize$\pm$0.03}} \\
        SL-R & 6.370{\scriptsize$\pm$0.756} & \textbf{0.72{\scriptsize$\pm$0.13}} & 0.50{\scriptsize$\pm$0.04} \\
        SVM & 0.439{\scriptsize$\pm$0.071} & 0.71{\scriptsize$\pm$0.13} & 0.46{\scriptsize$\pm$0.04} \\
        \bottomrule
    \end{tabular}
\end{table}

\subsection{CAV-based Explanation Methods}
\label{subsec:xai_methods}
After quantitatively comparing CAVs computed by our approach and the SVM-based approach, we investigate the qualitative performance with downstream concept-based explanation methods.
For this comparison, we substitute SVM-based CAV computation with \methodname and contrast the generated explanations, showcasing its ability to act as a drop-in replacement.
As concept-based explanation methods, we select Testing with Concept Activation Vectors (TCAV) \cite{kim2018interpretabilityfeatureattributionquantitative} and Automatic Concept-based Explanations (ACE) \cite{ghorbani2019towards}.

\paragraph{Testing with Concept Activation Vectors (TCAV)}
\begin{figure*}[!htb]
    \centering
    \includegraphics[width=\linewidth]{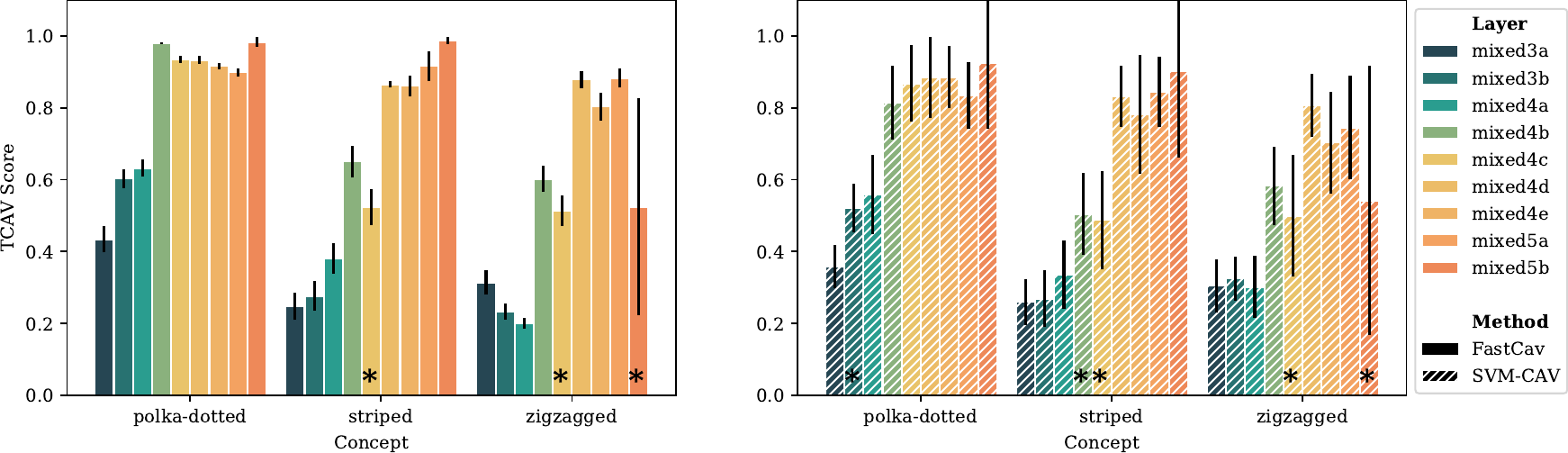}
    \caption{
    TCAV scores for various GoogleNet \cite{szegedy2015going} layers.
    We compare the concepts \concept{polka-dotted}, \concept{striped}, and \concept{zigzagged} for the class \emph{ladybug} using \methodname against \oldmethodname.
    We follow \cite{kim2018interpretabilityfeatureattributionquantitative} and mark CAVs that are not statistically significant with ``*''.
    }
    \label{fig:tcav:ladybug}
\end{figure*}

TCAV calculates the directional derivative of a selected class $k$ with respect to the activations of a layer $l$, i.e., $g_l(x)$.
For a given input $x$, the sensitivity score for a concept $c$ with corresponding CAV $v_c^l$ is $\nabla h_{l,k}(g_l(x)) \cdot v_c^l$.
The corresponding TCAV score is now the fraction of a set of inputs of the selected class that have positive sensitivity scores.
Intuitively, this scalar measures the importance of a concept in layer $l$ for the output node corresponding to class $k$.

We qualitatively evaluate \methodname by repeating an experiment of %
\cite{kim2018interpretabilityfeatureattributionquantitative}.
Specifically, we select all layers of GoogleNet \cite{szegedy2015going} trained on ImageNet \cite{russakovsky2015imagenet} and calculate the TCAV scores for various classes and concepts.
In \Cref{fig:tcav:ladybug}, we compare the results for \methodname against \oldmethodname for the class \emph{ladybug} and the concepts \concept{polka dotted}, \concept{striped}, and \concept{zigzagged}.
In \Cref{apx:tcav}, we include more examples.

The insights into the GoogleNet model are consistent between both our approach and the SVM-based method.
For \emph{ladybug}, the concepts \concept{polka dotted} and \concept{striped} are more important than \concept{zigzagged}.
This observation especially applies in later layers of the network.
Further, both methods agree that \concept{polka dotted} is the most important for earlier layers.
We hypothesize that the usage of \concept{striped} in later layers is due to blades of grass in many \emph{ladybug} images.

In contrast to the similar average results, we observe a clear distinction for the standard deviations of the TCAV score over multiple repetitions.
Here, \methodname leads in most cases to a smaller variance in the measured scores.
This observation confirms our intra-method robustness results in \Cref{tab:compareFastCavwithSVMCAV}.
Additionally, this result and the overall similarity between both \methodname and \oldmethodname hold for other qualitative examples in \Cref{apx:tcav}. %
Hence, we conclude that our approach leads to similar model explanations using TCAV as SVM-based CAV computation.

\paragraph{Automatic Concept-based Explanations (ACE)}

\begin{figure}[!tb]
    \begin{subfigure}{0.49\linewidth}
        \centering
        \includegraphics[width=\linewidth, trim=0cm 16cm 0cm 1cm, clip]{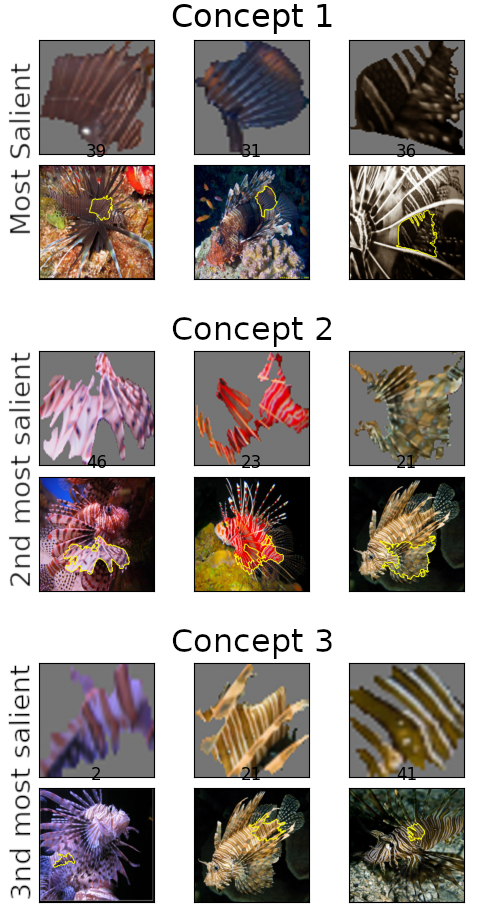}
        \caption{
        \methodname %
        }
        \label{fig:ace_lionfish_mean}
    \end{subfigure}
    \hfill
    \begin{subfigure}{0.49\linewidth}
        \centering
        \includegraphics[width=\linewidth, trim=0cm 16cm 0cm 1cm, clip]{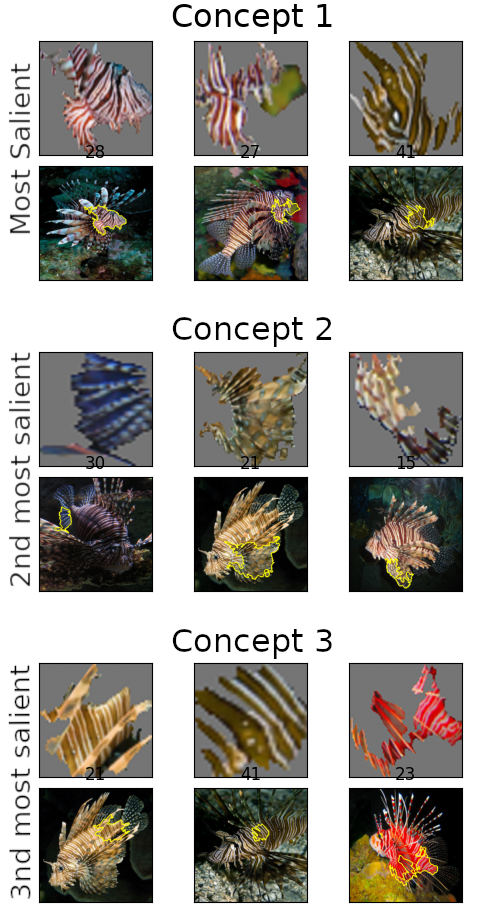}
        \caption{\oldmethodname %
        }
        \label{fig:ace_lionfish_cav}
    \end{subfigure}
    \vspace{-0.2cm}
    \caption{
    Most salient concepts discovered by ACE \cite{ghorbani2019towards} using either our \methodname or the established \oldmethodname.
    In both cases, we find the discovered patches containing stripes, which is congruent with the original observation in \cite{ghorbani2019towards}.
    }
    \label{fig:ace_lionfish_most_salient}
\end{figure}

\begin{figure}[!tb]
    \centering
    \begin{subfigure}{0.49\linewidth}
        \includegraphics[width=\linewidth]{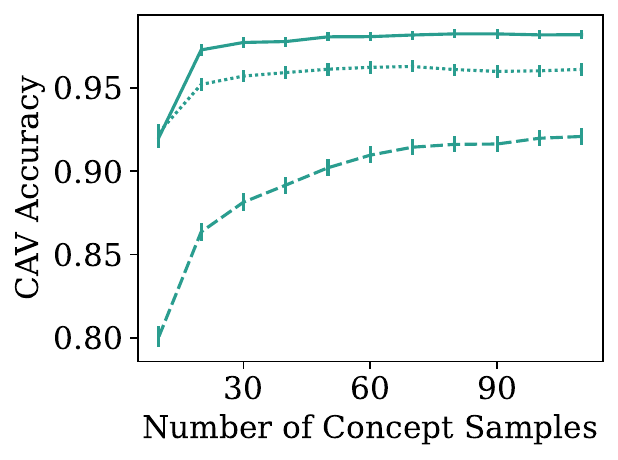}
        \caption{Impact of $|D_c|$.}
        \label{fig:abl:sample_size}
    \end{subfigure}
    \begin{subfigure}{0.49\linewidth}
        \includegraphics[width=\linewidth]{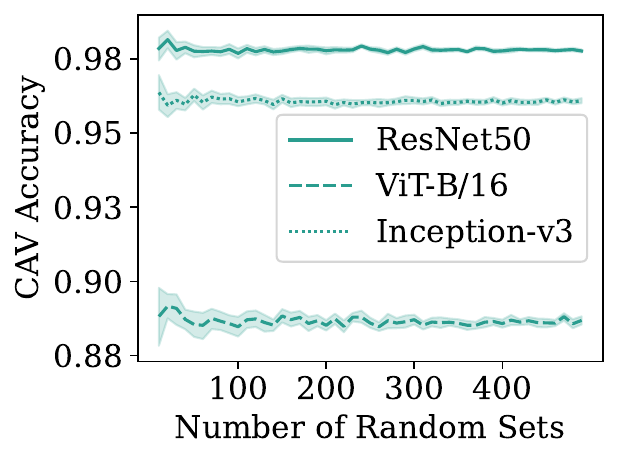}
        \caption{Number of resampled $D_r$.}
        \label{fig:abl:num_random_sets}
    \end{subfigure}
    
    \caption{Sensitivity analysis of \methodname to (a) the number of concept images and (b) number of resampled $D_r$. 
    Note the differences in y-axis scales.
    }
    \label{fig:abl}
\end{figure}

\begin{figure*}[!htb]
    \centering
    \includegraphics[width=\linewidth]{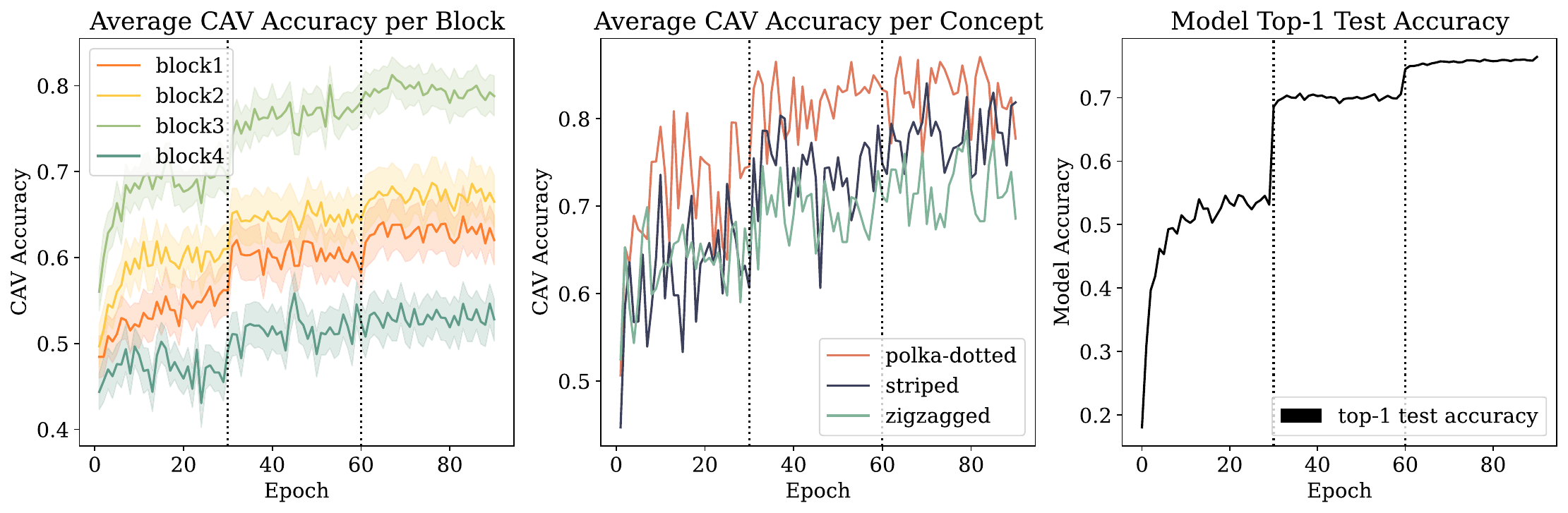}
    \caption{Evolution of various CAVs during training of a ResNet50 \cite{he2016deep} on ImageNet \cite{russakovsky2015imagenet}.
    On the left, we visualize the average accuracies achieved by CAVs after the final layers in each of the four ResNet blocks. %
    In the middle, we investigate CAVs for three specific concepts.
    On the right, we display the test accuracy during training.
    The vertical lines indicate epochs, after which the learning rate is divided by ten. 
    Note the differences in y-axis scales.
    }
    \label{fig:cavs_during_training}
\end{figure*}

ACE is based on TCAV and proposes to automatically discover and extract concepts that are important for a selected class.
In detail, they perform superpixel segmentation on a set of images from class $k$ using multiple resolutions to capture hierarchical concepts.
Next, these superpixels are grouped into concept sets using \cite{zhang2018unreasonable}.
The resulting sets are pruned and then ordered according to their importance for class $k$ using TCAV scores.

To evaluate \methodname for ACE qualitatively, we follow %
\cite{ghorbani2019towards} by analyzing the \texttt{mixed\_7a} layer in an Inception-v3  model \cite{szegedy2016rethinking} trained on ImageNet \cite{russakovsky2015imagenet}.

\Cref{fig:ace_lionfish_most_salient} visualizes the most salient concepts discovered with ACE. 
We include additional examples in \Cref{apx:ace}, following the class selection in \cite{ghorbani2019towards}.
In all cases, we find similar example patches for the discovered concepts.
In particular, the concept \concept{stripes} is important for the class \emph{lionfish} irrespective of whether we employ \methodname or \oldmethodname. 
Further, the TCAV scores for the most salient concepts are similar, with 0.78 for our approach and 0.73 for \oldmethodname.
These similarities between the discovered concepts and scores also apply to the other examples in \Cref{apx:ace}.
Hence, we conclude that \methodname is suitable for automatic concept discovery using ACE.

\subsection{\methodname Sensitivity Analysis}\label{subsec:ablationstudy}
Our approach to finding CAVs depends on the number of concept images, i.e., $|D_c|$ and $|D_r|$.
Hence, it is important to understand how many samples are necessary to compute robust concept directions.
Additionally, the robustness also depends on the number of random sets $D_r$ which are used for statistical testing,  \cite{kim2018interpretabilityfeatureattributionquantitative}.

Here, we conduct a sensitivity analysis to explore the influence of both variables. %
In particular, we vary the sizes of $D_c$ and $D_r$ as well as the number of available random sets.
Further, for this sensitivity analysis, we compute CAVs using \methodname for various networks.
Specifically, we use the pre-final layers of ResNet50 \cite{he2016deep}, ViT-16, \cite{dosovitskiy2020vit}, and Inception-v3 \cite{szegedy2016rethinking} pretrained on  ImageNet \cite{russakovsky2015imagenet}.

\paragraph{Influence of $|D_c|=|D_r|$}
We fix the total number of available random sets to 60 and vary the number of examples in the concept and random sets equally, evaluating the corresponding CAV accuracies. 
In \Cref{fig:abl:sample_size}, we visualize the influence of $|D_c|$ on the accuracy of the CAVs to distinguish between concept and random images. 
In particular, we find that for all three models, the performance remains high with more than 60 samples.
This indicates that with at least 60 examples, the CAV training process effectively captures the defining features of the concept.
Following this observation, we recommend using at least 60 examples per concept.

\vspace{-0.2cm}
\paragraph{Influence of the Number of Random Sets $D_r$}
Following the previous sensitivity analysis, we fix the number of examples in $D_c$ and $D_r$ to 60 and vary the number of random sets. %
In \Cref{fig:abl:num_random_sets}, we visualize the corresponding sensitivity of the CAV accuracies.
We see expected stabilization, i.e., lower variances, when using more random sets $D_r$.
However, the average accuracy remains approximately constant.
Hence, in our final experiment, we select 100 random sets. %

\subsection{Tracking CAVs During Training}
\label{subsec:training-analysis}
\paragraph{Setup} 
Using \methodname, we are able to track concepts during the training of a model with a high number of parameters. 
To show this, we train a ResNet50 \cite{he2016deep} on the ImageNet dataset \cite{russakovsky2015imagenet} from scratch. 
We follow the same setup as \cite{paszke2019pytorch} and include the full details in \Cref{appendix:tracking_cavs}. 
As concepts, we select the textures from \cite{bau2017network}.
Examples include \concept{polka-dotted}, \concept{striped}, and \concept{zigzagged}.
Specifically, we compute CAVs against 100 random sets after each training epoch and select the final layers of each of the four ResNet blocks.
Our analysis focuses on when and in which layer the model learns concepts during the training.

\begin{figure}[!htb]
    \centering
    \includegraphics[width=\linewidth]{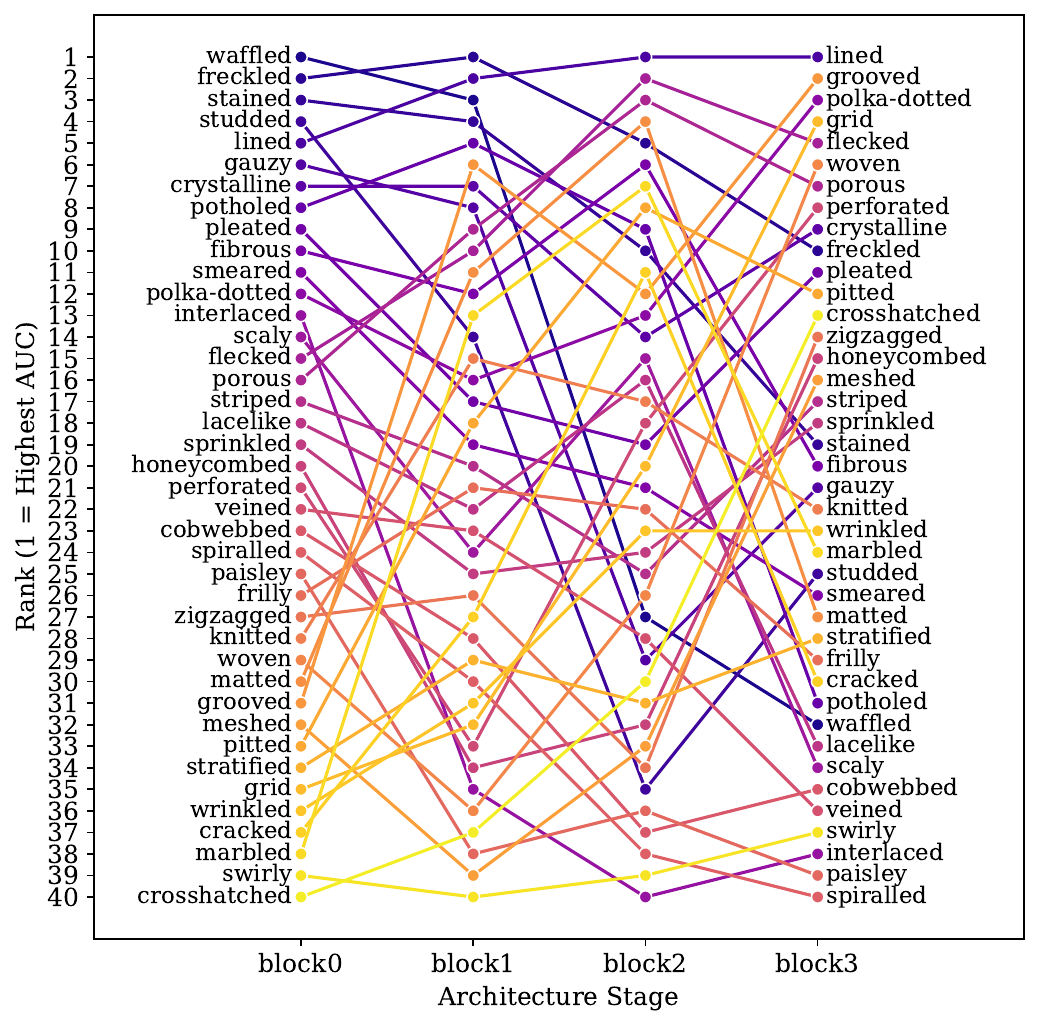}
    \caption{Evolution of various CAVs during training of a ResNet50 \cite{he2016deep} on ImageNet \cite{russakovsky2015imagenet}. We report the concept ranking of the achieved AUC scores during training for the different network blocks.
    }
    \label{fig:concept_ranking}
\end{figure}

\paragraph{Results} 
\Cref{fig:cavs_during_training} demonstrates that the model learns concepts during training, with visibly increasing CAV accuracy after each step. 
This aligns with improved model accuracy, suggesting that the model learns to employ relevant concepts for predictions.
We observe similar trends for specific concept examples, although results exhibit increased variability across the training steps compared to the average across concepts.
These results indicate that the performance of relevant CAVs correlates with overall model performance.

Notably, we observe stark increases in average CAV accuracy after each epoch, where the learning rate is divided by ten during training.
This observation aligns with abrupt increases in the top-1 test accuracy. 
Furthermore, we find that early and middle layers have a higher likelihood of learning textures than later layers. 
This result validates findings in other works like \cite{kim2018interpretabilityfeatureattributionquantitative,ghorbani2019towards,bau2017network}.
Our observations provide further evidence for this hypothesis and demonstrate that our approach can be used to study the learning dynamics of deep neural networks in a more fine-grained manner and for abstract concepts.

Lastly, we conduct an analysis ranking concepts by their area under the curve (AUC) during training and across layers.
Specifically, high AUC values indicate that a concept in a certain layer is learned early, and the CAV achieves sufficient accuracy during the complete training. 
We visualize the resulting rankings over the network blocks in \Cref{fig:concept_ranking} and observe notable changes in the learnability of concepts across layers.
For instance, concepts that rank highly in earlier network blocks may be less important and rank lower in later ones.
As a concrete example, consider the concept \concept{wave}, which is the highest-ranked concept in the first block.
However, its ranking drops to second after block two and ultimately deteriorates to 32nd in the final block. 
In contrast, the concept \concept{lined} remains among the highest-ranked concepts throughout all layers.
This observation underscores how the model prioritizes different concepts at various stages, with some foundational concepts giving way to more refined or abstract ones in deeper layers.

\section{Conclusion}
In this work, we introduced \methodname, a novel approach to efficiently compute Concept Activation Vectors (CAVs) for explaining deep neural networks. 
By leveraging ideas from superposition \cite{elhage2022toy}, our method accelerates CAV extraction.
Specifically, we take the normalized vector pointing towards the concept example mean in activation space, which is significantly faster than learning a linear SVM classifier.
We further provide a theoretical foundation for our method and give specific assumptions under which \methodname is equivalent to the established SVM-based approach.
Our empirical results demonstrate that \methodname achieves a significant reduction in computational cost, with speedups of up to \maxspeedup (and an average speedup of \avgspeedup), while maintaining at least similar CAV quality. 
We further investigate the variations in downstream performances using TCAV \cite{kim2018interpretabilityfeatureattributionquantitative} and ACE \cite{ghorbani2019towards}, and find only marginal differences in the results.
To showcase the capabilities of \methodname, we apply it to analyze the training of a model, tracking the evolution of concepts during training and across network layers. 
Our analysis validates the notion that textural concepts are learned in the middle layers of a network, aligning with prior works \cite{bau2017network,kim2018interpretabilityfeatureattributionquantitative,ghorbani2019towards}.

Overall, \methodname serves as a drop-in replacement to calculate CAVs, which leads to lower run times and maintains high quality.
While we thoroughly evaluate the performance of the generated CAVs, we acknowledge that further analysis of other CAV properties, such as locality, consistency, and entanglement\cite{nicolson2024explainingexplainabilityunderstandingconcept}, is an interesting direction for future work.
We hope that the gain in computational speed of \methodname facilitates concept-based explanations that were previously infeasible.

\section*{Impact Statement}
This paper presents work whose goal is to advance the field of Machine Learning, specifically, explainability. 
There are many potential societal consequences of our work, none of which we feel must be specifically highlighted here.

\nocite{langley00}

\bibliography{main.bib}
\bibliographystyle{icml2025}

\newpage
\appendix
\onecolumn

\section{\methodname --- Additional Details}
\label{apx:theory}
In this section, we provide additional theoretical details connected to \Cref{sec:method} in our main paper.

\subsection{Expectation of the Global Mean Under Gaussian Mixture Assumption}
\label{apx:expectation}

In \methodname, we utilize the maximum likelihood estimator for the mean of a sample.
This estimator is generally unbiased under \emph{i.i.d.} samples following a Gaussian distribution, e.g., \cite{bishop2006pattern}.
In \Cref{eq:mlest} in our main paper, we apply this to the global mean over $D_c \cup D_r$.
However, consider the following related estimators:
\begin{equation}
    \hat{\mu}_{D_c} = \frac{1}{|D_c|} \sum_{x \in D_c} g_l(x) \text{,\hspace{0.5cm}and\hspace{0.5cm} }\hat{\mu}_{D_r} = \frac{1}{|D_r|} \sum_{x \in D_r} g_l(x).
\end{equation}
We now assume that both $D_c$ and $D_r$ contain \emph{i.i.d.} samples so that the corresponding activations $g_l(x)$ follow two respective Gaussian distributions $\mathcal{N}(\mu_c, \Sigma_c)$ and $\mathcal{N}(\mu_r, \Sigma_r)$.
In other words, we assume that both the random sample and concept example activations follow independent Gaussian distributions in the activation space of $g_l$.
Then from the unbiasedness of the maximum likelihood estimator \cite{bishop2006pattern} follows both
\begin{equation}
    \mathbb{E}[\hat{\mu}_{D_c}] = \mu_c \text{,\hspace{0.5cm}and\hspace{0.5cm} } \mathbb{E}[\hat{\mu}_{D_r}] = \mu_r.
\end{equation}

Using these equations, we now study the expectation of the global mean estimator $\hat{\mu}_{D_c\cup D_r}$.
In \cite{kim2018interpretabilityfeatureattributionquantitative}, both the random sample and the set of concept examples are of equal size, which we follow in our work.
Hence, we assume a uniform mixture of the two Gaussians specified above.
In other words, the mixture weights are $\nicefrac{1}{2}$, and $|D_c| = |D_r|$.
Then the expectation of \Cref{eq:mlest} becomes 
\begin{align} 
    \mathbb{E}[\hat{\mu}_{D_c\cup D_r}] &= \mathbb{E}\left[\frac{1}{|D_c\cup D_r|} \sum_{x \in D_c\cup D_r} g_l(x)\right]\\
    &= \mathbb{E}\left[\frac{1}{|D_c| + |D_r|} \sum_{x \in D_c} g_l(x) + \frac{1}{|D_c| + |D_r|} \sum_{x \in D_r} g_l(x)\right]\\
    &= \mathbb{E}\left[\frac{1}{2}\frac{1}{|D_c|} \sum_{x \in D_c} g_l(x) + \frac{1}{2}\frac{1}{|D_r|} \sum_{x \in D_r} g_l(x)\right]\\
    &= \frac{1}{2} \left(\mathbb{E}\left[\frac{1}{|D_c|} \sum_{x \in D_c} g_l(x)\right] + \mathbb{E}\left[\frac{1}{|D_r|} \sum_{x \in D_r} g_l(x)\right]\right)\\
    &= \frac{1}{2}\left(\mathbb{E}[\hat{\mu}_{D_c}] + \mathbb{E}[\hat{\mu}_{D_r}]\right)\\
    &= \frac{\mu_c + \mu_r}{2}.
\end{align}
Intuitively, the global mean is the middle point between the mean of the random samples and the concept examples in the activation space of $g_l$.
We then use this point to normalize the activations before calculating the direction towards the concept mean as our CAV $v_c^l$.
This vector is under the specified assumptions and an expectation over the sample $D_c \cup D_r$ proportional to \Cref{eq:cav-exp} in our main paper.

\subsection{Empirical Ratio of Support Vectors}
\label{apx:svs}

\begin{table}[H]
    \centering
    \caption{Percentage of support vectors of SVMs that learned the representation of 40 different concepts \cite{bau2017network}. 
    Each SVM was trained to separate 50 example concept images from 50 random images.
    We average over 30 redrawn $D_r$ and split our report into the two classes defined by the sets $D_c$ and $D_r$.
    }
    \label{tab:concept_support_vectors}
    \begin{tabular}{lccc}
        \toprule
        && \multicolumn{2}{c}{Support Vector Percentage}\\
        \cmidrule{3-4}
        Model Architecture & $\varnothing$ Activation Dimensionality  & $D_c$ & $D_r$ \\
        \midrule
        ResNet50 \cite{he2016deep}      & 100,352    & 77.05\% & 98.17\% \\
        ConvNeXt-XXL \cite{liu2022convnet}  & 196,608    & 52.31\% & 90.40\% \\
        RegNetY \cite{singh2022revisiting}      & 1,064,448  &  78.00\% & 97.66\% \\

        ViT-B/16 \cite{dosovitskiy2020vit}     & 151,296    & 73.27\% & 78.87\% \\
        ViT-H/14-CLIP \cite{dosovitskiy2020vit} & 328,960    & 87.85\% & 88.72\% \\
        
        EVA-G/14 \cite{fang2023eva}     & 812,416    & 78.69\% & 90.60\% \\
        EVA-02-L/14 \cite{fang2024eva}   & 1,049,600  & 88.45\% & 99.67\% \\
        \bottomrule
    \end{tabular}
\end{table}

In \Cref{tab:concept_support_vectors}, we list the percentages of support vectors when using an SVM to compute CAVs in the pre-final layer of various networks.
In all cases, we observe that the majority of activation vectors are support vectors.
This observation is congruent with other works that study this behavior for high dimensionalities in more detail, e.g., \cite{muthukumar2021classification,hsu2021proliferation}.
However, interestingly, we find that the percentages are higher for the sets of random images $D_r$ in comparison to concept images $D_c$.
We hypothesize that this might be a consequence of more compact activations for images sharing a semantic concept.

\subsection{Runtime Benchmark of SVM Implementations}
\label{apx:svm-speed}

As described in our main paper, we benchmark the speed of three common implementations of linear SVMs.
First, the SMO \cite{platt1998fast} implementation contained in \cite{pedregosa2011scikit}, which relies on \cite{chang2011libsvm}.
Second, we use the solution for the primal problem implemented in \cite{fan2008liblinear}, which is also included in \cite{pedregosa2011scikit}.
Lastly, using Hinge loss, it is possible to approximate a linear SVM classifier using stochastic gradient descent (SGD), e.g., \cite{pedregosa2011scikit}.
We visualize the measured achieved runtimes in \Cref{fig:appendix:svm_benchmark} for both increasing numbers of examples $n$ and increasing input dimensions $d$.
\begin{figure*}[!htb]
    \begin{subfigure}[t]{0.49\textwidth}
        \centering
        \includegraphics[width=\linewidth]{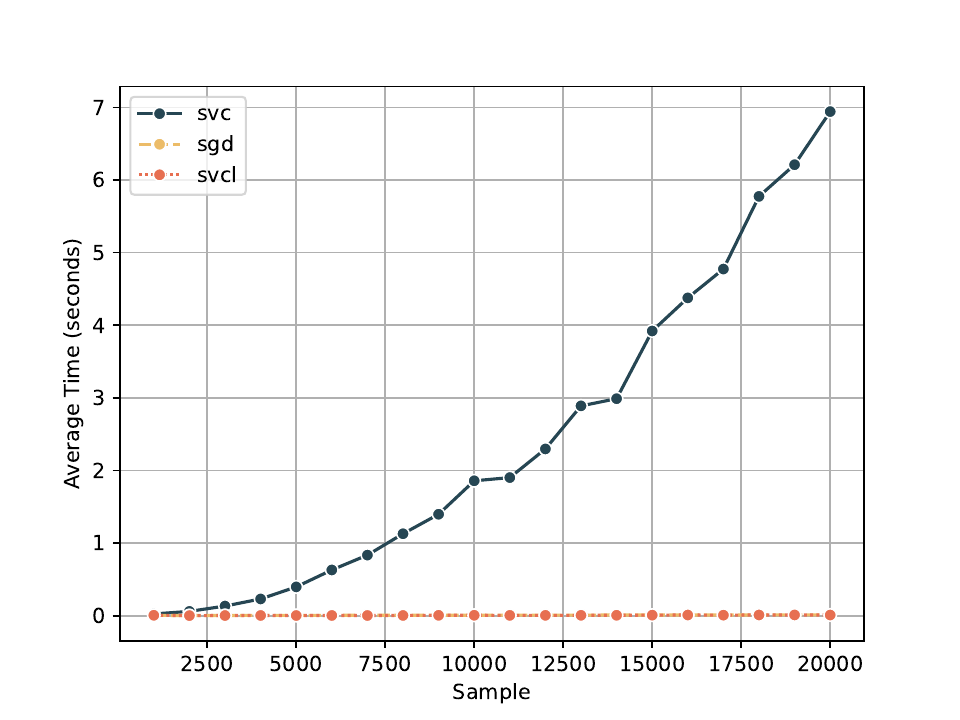}
        \caption{Dependency on the number of samples $n$. We fix the dimensionality to $d=2$.}
        \label{fig:appendix:svm_samples}
    \end{subfigure}
    \hfill
    \begin{subfigure}[t]{0.49\textwidth}
        \centering
        \includegraphics[width=\linewidth]{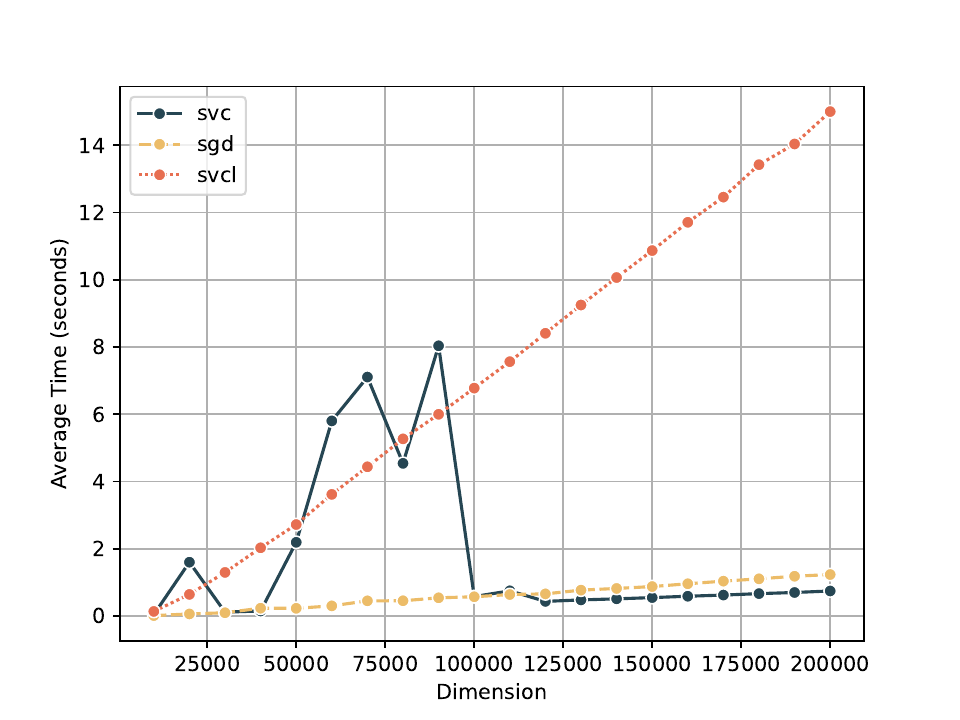}
        \caption{
        Dependency on dimensionality $d$. We set the total number of samples to $n=50$.
        }
        \label{fig:appendix:svm_dimension}
    \end{subfigure}
    \vspace{-0.2cm}
    \caption{
    It can be seen that our method results in similar results.
    }
    \label{fig:appendix:svm_benchmark}
\end{figure*}
For our application, we find that the SGD-based optimization performs best.
Hence, we use this implementation for the classical SVM-based CAV calculation in our experiments.
Note that this also follows related work \cite{kim2018interpretabilityfeatureattributionquantitative}, which uses the same implementation.

\section{Experiments}

\subsection{Experimental Setup Details}
\label{apx:setup}
\paragraph{Technical details}
Our experimental setup utilized a system equipped with two Intel Xeon Platinum 8260 processors (48 cores total) and 384 GiB of RAM, along with 8 NVIDIA Tesla V100 GPUs, each providing 32 GiB of memory. 
For the calculation of the CAVs, we exclusively used the CPU-based infrastructure to ensure a consistent and comparable environment for evaluating both \methodname and \oldmethodname. 
To train the ResNet50, we leveraged the full capabilities of the system.

\paragraph{Models}
For our experiments, we either report the metrics with respect to the last layer mentioned in the list or we average over all layers.
In the case of architectures composed of repeated blocks, e.g., ResNet, we refer in the list to the final layer of each block.
The notation used follows from the implementations in \cite{rw2019timm} or \cite{wolf2019huggingface} to ensure reproducibility.
\begin{itemize}
    \item ResNet50 \cite{he2016deep}: conv1, layer1, layer2, layer3, layer4
    \item Inception-v3 \cite{szegedy2016rethinking}: Mixed\_5b, Mixed\_5c, Mixed\_5d, Mixed\_6a, Mixed\_6b, Mixed\_6c, Mixed\_6d, Mixed\_6e, Mixed\_7a, Mixed\_7b, Mixed\_7c
    \item ConvNeXt-XXLarge \cite{liu2022convnet}: stem, stage1, stage2, stage3, stage4
    \item RegNetY \cite{singh2022revisiting}: stem, stage1, stage2, stage3, stage4
    \item ViT-B/16 \cite{dosovitskiy2020vit}: embeddings, encoder.block.0, encoder.block.1, encoder.block.2, encoder.block.3, encoder.block.4, encoder.block.5, encoder.block.6, encoder.block.7, encoder.block.8, encoder.block.9, encoder.block.10, encoder.block.11
    \item ViT-H/14-CLIP \cite{dosovitskiy2020vit,radford2021learning}: patch\_embed, blocks.0, blocks.7, blocks.15, blocks.23, blocks.31
    \item EVA-G/14 \cite{fang2023eva}: patch\_embedding, encoder.block.0, encoder.block.9, encoder.block.19, encoder.block.29, encoder.block.39
    \item EVA-02-L/14 \cite{fang2024eva}: patch\_embedding, encoder.block.0, encoder.block.5, encoder.block.11, encoder.block.17, encoder.block.23
\end{itemize}

\paragraph{Setup of $D_c$ and $D_r$}
For the experiments in \Cref{subsec:comparison}, we utilize the Broden dataset \cite{bau2017network} to define the concept sets $D_c$. 
Broden describes each of its 48 pattern-based concepts with 120 images. 
Here, we sample 60 concept images per set $D_c$ from those 120 images once.
Similarly, we sample 60 random images from ImageNet \cite{russakovsky2015imagenet} to define one set $D_r$  following \cite{kim2018interpretabilityfeatureattributionquantitative}. 

Later, to calculate the accuracy of a computed CAV (see \Cref{eq:class}), we sample independent validation data from the remaining 60 concept examples of the Broden dataset, \cite{bau2017network}, and ImageNet \cite{russakovsky2015imagenet}, for $D_c$ and $D_r$, respectively.

\subsection{Additional Empirical Comparisons of CAV Computation}
\label{apx:exp1}

\begin{figure}[H]
    \centering
    \includegraphics[width=0.5\linewidth]{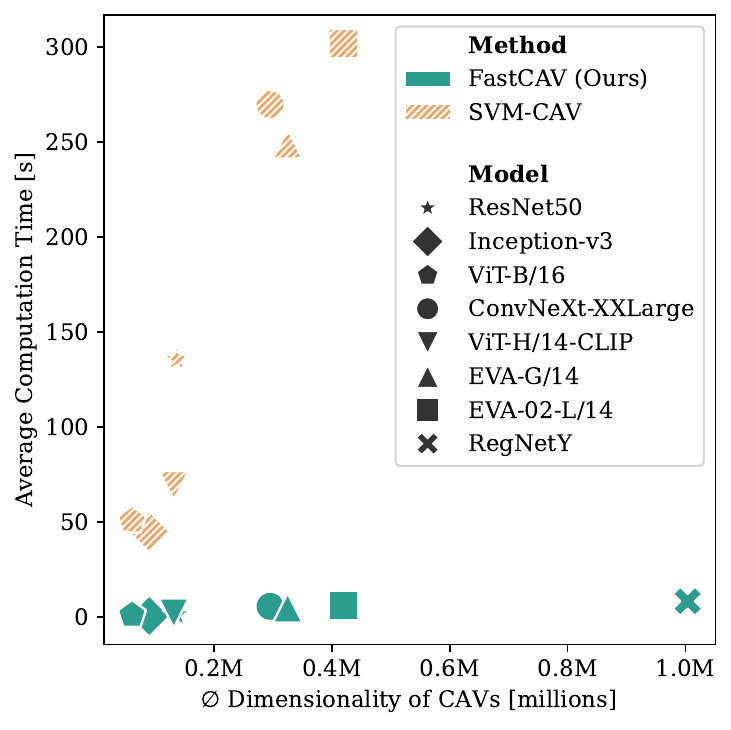}
    
    \caption{
    Comparison of computational efficiency between \methodname and the established \oldmethodname for CAVs across different models for multiple layers.
    The average time taken to calculate a CAV for each method is plotted against the average dimensionality of the activation spaces, demonstrating the significant speedup achieved by our \methodname. 
    The differences in computation time between \methodname and the \oldmethodname are statistically significant ($p < 0.01$) for all models. 
    Note that \oldmethodname for the model RegNetY exceeded our maximum computational time.}
    \label{fig:apx:teaser}
\end{figure}

As a complement to \Cref{fig:introfigure} in our main paper, we include \Cref{fig:apx:teaser}, where we show the timings for all models averaged over the network layers.
Note the clear difference for the RegNetY model \cite{singh2022revisiting} compared to \Cref{fig:introfigure}, due to the increased average activation space dimensionality.
Calculating CAVs using SVMs was not feasible in our setup.

\begin{table*}[!htb] %
    \small
    \centering
    \caption{Comparing our approach \methodname with \oldmethodname.
    In this table, we focus only on the pre-final layer before the prediction, which should encode the most semantic information.
    \textbf{Bold values} indicate better results.
    ``N/A'' indicates that no results were produced due to the overall computational time for the respective model exceeding four days.}
    \label{tab:lastlayercompareFastCavwithSVMCAV}
    \begin{tabular}{lcrlrlrl:c}
        \toprule
          & & \multicolumn{2}{c}{ Comp. Time [$s$] $\downarrow$}  &\multicolumn{2}{c}{Accuracy $\uparrow$} & \multicolumn{3}{c}{Similarity $\uparrow$}\\
          \cmidrule(lr){3-4} \cmidrule(lr){5-6} \cmidrule(lr){7-9}
          Model & $\varnothing$ Dim. &  \methodname & \oldmethodname & \methodname  & \oldmethodname &  \methodname  & \oldmethodname & Inter-Method \\
        \midrule
        
        Inception-v3 & 131,072 & \textbf{0.3{\scriptsize $\pm$0.05}} & 17.3{\scriptsize $\pm$2.19} & \textbf{0.95{\scriptsize $\pm$0.06}} & \textbf{0.95{\scriptsize $\pm$0.05}}  & \textbf{0.796{\scriptsize $\pm$0.016}} & 0.300{\scriptsize $\pm$0.050} & 0.823{\scriptsize $\pm$0.039}  \\
        ResNet50 & 100,352& \textbf{0.3{\scriptsize $\pm$0.84}} & 13.8{\scriptsize $\pm$2.04}  & \textbf{0.98{\scriptsize $\pm$0.04}} & \textbf{0.98{\scriptsize $\pm$0.04}} & \textbf{0.823{\scriptsize $\pm$0.010}} & 0.190{\scriptsize $\pm$0.057} & 0.775{\scriptsize $\pm$0.045}  \\
        ConvNeXt-XXL & 196,608 & \textbf{1.4{\scriptsize $\pm$0.10}} & 45.7{\scriptsize $\pm$4.73} & 0.96{\scriptsize $\pm$0.05} & \textbf{0.98{\scriptsize $\pm$0.04}}  & \textbf{0.859{\scriptsize $\pm$0.031}} & 0.218{\scriptsize $\pm$0.059} & 0.816{\scriptsize $\pm$0.042} \\
        RegNetY & 1,064,448 & \textbf{6.1{\scriptsize $\pm$1.04}} & 312.4{\scriptsize $\pm$63.23} & 0.97{\scriptsize $\pm$0.05}  & \textbf{0.98{\scriptsize $\pm$0.03}}  &  \textbf{0.825{\scriptsize $\pm$0.009}} & 0.128{\scriptsize $\pm$0.066} & 0.675{\scriptsize $\pm$0.065} \vspace{0.15cm} \\
        ViT-B/16 & 151,296 & \textbf{1.0{\scriptsize $\pm$0.03}} & 50.5{\scriptsize $\pm$5.78} & \textbf{0.87{\scriptsize $\pm$0.09}} & 0.85{\scriptsize $\pm$0.09}  & \textbf{0.813{\scriptsize $\pm$0.022}} & 0.454{\scriptsize $\pm$0.041} & 0.854{\scriptsize $\pm$0.048} \\
        ViT-H/14-CLIP & 328,960 & \textbf{1.8{\scriptsize $\pm$0.07}} & 67.0{\scriptsize $\pm$9.67} & \textbf{0.94{\scriptsize $\pm$0.06}} & 0.92{\scriptsize $\pm$0.07}  & \textbf{0.654{\scriptsize $\pm$0.080}} & 0.328{\scriptsize $\pm$0.084} & 0.894{\scriptsize $\pm$0.066} \\
        EVA-G/14 & 812,416 & \textbf{4.7{\scriptsize $\pm$0.12}} & 282.8{\scriptsize $\pm$32.43} & \textbf{0.95{\scriptsize $\pm$0.04}} & 0.94{\scriptsize $\pm$0.05}  & \textbf{0.762{\scriptsize $\pm$0.059}} & 0.201{\scriptsize $\pm$0.066} & 0.848{\scriptsize $\pm$0.044} \\
        EVA-02-L/14 & 1,049,600 & \textbf{6.1{\scriptsize $\pm$0.14}} & 252.2{\scriptsize $\pm$29.57} & 0.93{\scriptsize $\pm$0.07} & \textbf{0.95{\scriptsize $\pm$0.05}} & \textbf{0.637{\scriptsize $\pm$0.075}} & 0.119{\scriptsize $\pm$0.065} & 0.747{\scriptsize $\pm$0.059}   \\
        
        \bottomrule
    \end{tabular}
\end{table*}

Similarly, we provide \Cref{tab:lastlayercompareFastCavwithSVMCAV} an equivalent of \Cref{tab:compareFastCavwithSVMCAV} containing the detailed numbers for only the pre-final layers of all networks in our study.

Both \Cref{fig:apx:teaser} and \Cref{tab:lastlayercompareFastCavwithSVMCAV} confirm the observations discussed in our main paper.
However, as an additional explorative analysis, we investigate cases where \methodname and SVM-based computation differ significantly.

\subsubsection{Comparison Against Other CAV Computation Approaches}
\label{apx:add-comp}

To provide further evidence for our claims regarding the improved computation speed, we rerun a smaller version with 30 CAVs per concept of our first experiment with additional computation methods (compare to \Cref{tab:compareFastCavwithSVMCAV}).
Specifcally, we compare against the following approaches: logistic regression (LR) \cite{pfau2021robust}, sparsified logistic regression (S-LR) \cite{singla2021using}, Ridge classification as a faster alternative \cite{pedregosa2011scikit}, and classical LDA, which is closely related to our approach (see \Cref{sec:method}).

\begin{table*}[!htb] %
    \small
    \centering
    \caption{Comparing our approach \methodname with other CAV calculation methods regarding computation time.
    \textbf{Bold values} indicate better results.
    }
    \label{tab:alllayercompareFastCavwithOtherTime}
    \begin{tabular}{lcccccc}
    \toprule
     & \multicolumn{6}{c}{Comp. Time [$s$] $\downarrow$} \\
    \cmidrule(lr){2-7}
    Model & FastCAV (Ours) & SVM & LDA & LR & S-LR & Ridge \\
    \midrule
    ConvNeXt-XXLarge
      & \textbf{{0.019}{\scriptsize $\pm$ 0.013}}  & {1.167}{\scriptsize $\pm$ 0.795}  & {16.842}{\scriptsize $\pm$ 11.911}  & {9.360}{\scriptsize $\pm$ 2.547}  & {9.549}{\scriptsize $\pm$ 2.734}  & {0.695}{\scriptsize $\pm$ 0.478} \\
    EVA-02-L/14
      & \textbf{{0.024}{\scriptsize $\pm$ 0.001}}  & {1.524}{\scriptsize $\pm$ 0.594}  & {23.469}{\scriptsize $\pm$ 3.404}  & {7.659}{\scriptsize $\pm$ 4.693}  & {7.693}{\scriptsize $\pm$ 4.731}  & {0.923}{\scriptsize $\pm$ 0.019} \\
    EVA-G/14
      & \textbf{{0.018}{\scriptsize $\pm$ 0.001}}  & {1.109}{\scriptsize $\pm$ 0.263}  & {19.135}{\scriptsize $\pm$ 2.200}  & {8.499}{\scriptsize $\pm$ 4.763}  & {8.457}{\scriptsize $\pm$ 4.696}  & {0.681}{\scriptsize $\pm$ 0.023} \\
    Inception-v3
      & \textbf{{0.013}{\scriptsize $\pm$ 0.020}}  & {1.366}{\scriptsize $\pm$ 0.943}  & {10.755}{\scriptsize $\pm$ 7.188}  & {8.681}{\scriptsize $\pm$ 6.076}  & {8.648}{\scriptsize $\pm$ 5.960}  & {0.409}{\scriptsize $\pm$ 0.258} \\
    RegNetY
      & \textbf{{0.039}{\scriptsize $\pm$ 0.017}}  & {5.207}{\scriptsize $\pm$ 1.054}  & {54.640}{\scriptsize $\pm$ 37.392}  & {11.529}{\scriptsize $\pm$ 4.791}  & {11.747}{\scriptsize $\pm$ 4.917}  & {2.184}{\scriptsize $\pm$ 1.300} \\
    ResNet50
      & \textbf{{0.016}{\scriptsize $\pm$ 0.019}}  & {2.908}{\scriptsize $\pm$ 2.440}  & {7.827}{\scriptsize $\pm$ 5.601}  & {8.375}{\scriptsize $\pm$ 5.509}  & {8.474}{\scriptsize $\pm$ 5.687}  & {0.314}{\scriptsize $\pm$ 0.225} \\
    ViT-B/16
      & \textbf{{0.005}{\scriptsize $\pm$ 0.002}}  & {0.852}{\scriptsize $\pm$ 0.134}  & {3.421}{\scriptsize $\pm$ 0.443}  & {6.517}{\scriptsize $\pm$ 2.907}  & {6.506}{\scriptsize $\pm$ 2.909}  & {0.145}{\scriptsize $\pm$ 0.016} \\
    ViT-H/14-CLIP
      & \textbf{{0.008}{\scriptsize $\pm$ 0.001}}  & {0.456}{\scriptsize $\pm$ 0.078}  & {8.867}{\scriptsize $\pm$ 1.037}  & {7.091}{\scriptsize $\pm$ 2.658}  & {7.118}{\scriptsize $\pm$ 2.647}  & {0.288}{\scriptsize $\pm$ 0.013} \\
    \bottomrule
    \end{tabular}
\end{table*}

\begin{table*}[!htb] %
    \small
    \centering
    \caption{Comparing our approach \methodname with other CAV calculation methods regarding accuracy.
    \textbf{Bold values} indicate better results.
    }
    \label{tab:alllayercompareFastCavwithOtherAcc}
    \begin{tabular}{lcccccc}
    \toprule
     & \multicolumn{6}{c}{Accuracy $\uparrow$} \\
    \cmidrule(lr){2-7}
    Model & FastCAV (Ours) & SVM & LDA & LR & S-LR & Ridge \\
    \midrule
    ConvNeXt-XXLarge
      & {0.92}{\scriptsize $\pm$ 0.08}  & {0.95}{\scriptsize $\pm$ 0.06}  & {0.90}{\scriptsize $\pm$ 0.15}  & \textbf{{0.96}{\scriptsize $\pm$ 0.05}}  & {0.96}{\scriptsize $\pm$ 0.05}  & {0.96}{\scriptsize $\pm$ 0.05} \\
    EVA-02-L/14
      & {0.90}{\scriptsize $\pm$ 0.14}  & {0.90}{\scriptsize $\pm$ 0.15}  & {0.79}{\scriptsize $\pm$ 0.20}  & {0.91}{\scriptsize $\pm$ 0.15}  & {0.91}{\scriptsize $\pm$ 0.15}  & \textbf{{0.91}{\scriptsize $\pm$ 0.15}} \\
    EVA-G/14
      & {0.89}{\scriptsize $\pm$ 0.15}  & {0.88}{\scriptsize $\pm$ 0.15}  & {0.81}{\scriptsize $\pm$ 0.21}  & \textbf{{0.90}{\scriptsize $\pm$ 0.15}}  & {0.90}{\scriptsize $\pm$ 0.15}  & {0.90}{\scriptsize $\pm$ 0.15} \\
    Inception-v3
      & \textbf{{0.95}{\scriptsize $\pm$ 0.06}}  & {0.93}{\scriptsize $\pm$ 0.08}  & {0.76}{\scriptsize $\pm$ 0.19}  & {0.89}{\scriptsize $\pm$ 0.14}  & {0.89}{\scriptsize $\pm$ 0.14}  & {0.89}{\scriptsize $\pm$ 0.15} \\
    RegNetY
      & {0.97}{\scriptsize $\pm$ 0.05}  & \textbf{{0.98}{\scriptsize $\pm$ 0.03}}  & {0.63}{\scriptsize $\pm$ 0.15}  & {0.96}{\scriptsize $\pm$ 0.05}  & {0.96}{\scriptsize $\pm$ 0.05}  & {0.98}{\scriptsize $\pm$ 0.04} \\
    ResNet50
      & {0.89}{\scriptsize $\pm$ 0.15}  & {0.87}{\scriptsize $\pm$ 0.15}  & {0.72}{\scriptsize $\pm$ 0.20}  & {0.92}{\scriptsize $\pm$ 0.11}  & {0.92}{\scriptsize $\pm$ 0.11}  & \textbf{{0.93}{\scriptsize $\pm$ 0.12}} \\
    ViT-B/16
      & {0.83}{\scriptsize $\pm$ 0.13}  & {0.81}{\scriptsize $\pm$ 0.14}  & \textbf{{0.89}{\scriptsize $\pm$ 0.12}}  & {0.82}{\scriptsize $\pm$ 0.14}  & {0.82}{\scriptsize $\pm$ 0.14}  & {0.82}{\scriptsize $\pm$ 0.15} \\
    ViT-H/14-CLIP
      & {0.88}{\scriptsize $\pm$ 0.15}  & {0.87}{\scriptsize $\pm$ 0.15}  & {0.82}{\scriptsize $\pm$ 0.17}  & {0.89}{\scriptsize $\pm$ 0.15}  & \textbf{{0.89}{\scriptsize $\pm$ 0.15}}  & {0.89}{\scriptsize $\pm$ 0.16} \\
    \bottomrule
    \end{tabular}
\end{table*}

\begin{table*}[!htb] %
    \small
    \centering
    \caption{Comparing our approach \methodname with other CAV calculation methods regarding similarity.
    \textbf{Bold values} indicate better results.
    }
    \label{tab:alllayercompareFastCavwithOtherSim}
    \begin{tabular}{lcccccc}
    \toprule
     & \multicolumn{6}{c}{Similarity $\uparrow$} \\
    \cmidrule(lr){2-7}
    Model & FastCAV (Ours) & SVM & LDA & LR & S-LR & Ridge \\
    \midrule
    ConvNeXt-XXLarge
      & \textbf{{0.913}{\scriptsize $\pm$ 0.014}}  & {0.525}{\scriptsize $\pm$ 0.049}  & {0.521}{\scriptsize $\pm$ 0.105}  & {0.672}{\scriptsize $\pm$ 0.044}  & {0.671}{\scriptsize $\pm$ 0.044}  & {0.570}{\scriptsize $\pm$ 0.033} \\
    EVA-02-L/14
      & \textbf{{0.814}{\scriptsize $\pm$ 0.020}}  & {0.291}{\scriptsize $\pm$ 0.054}  & {0.443}{\scriptsize $\pm$ 0.086}  & {0.685}{\scriptsize $\pm$ 0.052}  & {0.688}{\scriptsize $\pm$ 0.050}  & {0.606}{\scriptsize $\pm$ 0.023} \\
    EVA-G/14
      & \textbf{{0.789}{\scriptsize $\pm$ 0.018}}  & {0.331}{\scriptsize $\pm$ 0.049}  & {0.418}{\scriptsize $\pm$ 0.070}  & {0.672}{\scriptsize $\pm$ 0.046}  & {0.670}{\scriptsize $\pm$ 0.050}  & {0.608}{\scriptsize $\pm$ 0.025} \\
    Inception-v3
      & \textbf{{0.826}{\scriptsize $\pm$ 0.011}}  & {0.387}{\scriptsize $\pm$ 0.058}  & {0.132}{\scriptsize $\pm$ 0.131}  & {0.602}{\scriptsize $\pm$ 0.037}  & {0.603}{\scriptsize $\pm$ 0.036}  & {0.562}{\scriptsize $\pm$ 0.026} \\
    RegNetY
      & \textbf{{0.825}{\scriptsize $\pm$ 0.009}}  & {0.128}{\scriptsize $\pm$ 0.066}  & {0.027}{\scriptsize $\pm$ 0.115}  & {0.697}{\scriptsize $\pm$ 0.063}  & {0.698}{\scriptsize $\pm$ 0.062}  & {0.675}{\scriptsize $\pm$ 0.020} \\
    ResNet50
      & \textbf{{0.791}{\scriptsize $\pm$ 0.030}}  & {0.398}{\scriptsize $\pm$ 0.062}  & {0.077}{\scriptsize $\pm$ 0.159}  & {0.652}{\scriptsize $\pm$ 0.042}  & {0.654}{\scriptsize $\pm$ 0.040}  & {0.602}{\scriptsize $\pm$ 0.026} \\
    ViT-B/16
      & \textbf{{0.752}{\scriptsize $\pm$ 0.029}}  & {0.394}{\scriptsize $\pm$ 0.048}  & {0.556}{\scriptsize $\pm$ 0.121}  & {0.592}{\scriptsize $\pm$ 0.058}  & {0.591}{\scriptsize $\pm$ 0.059}  & {0.508}{\scriptsize $\pm$ 0.037} \\
    ViT-H/14-CLIP
      & \textbf{{0.749}{\scriptsize $\pm$ 0.020}}  & {0.342}{\scriptsize $\pm$ 0.048}  & {0.551}{\scriptsize $\pm$ 0.149}  & {0.621}{\scriptsize $\pm$ 0.025}  & {0.621}{\scriptsize $\pm$ 0.025}  & {0.595}{\scriptsize $\pm$ 0.023} \\
    \bottomrule
    \end{tabular}
\end{table*}

We summarize the comparison results in \Cref{tab:alllayercompareFastCavwithOtherTime,tab:alllayercompareFastCavwithOtherAcc,tab:alllayercompareFastCavwithOtherSim}.
While all approaches deliver high accuracies, the logistic regression-based approaches achieve the second highest similarities after \methodname, which coincides with increased computation times.
In contrast, ridge classification offers a notable improvement in speed over SVM-based computation.
However, our \methodname still yields a substantial speed-up, while maintaining competitive performance to the more sophisticated approaches. 
Further, we can see that vanilla LDA is not enough to achieve similar gains. 
Hence, we argue that the assumptions we make in \Cref{subsec:linclass} are necessary under practical considerations.

\subsubsection{Significant Accuracy Differences Between \methodname and SVM-CAV}
\label{apx:differences}

While we believe that our empirical evaluation provides evidence for the practicality of our approach, the strong, but necessary, assumptions (see \Cref{subsec:linclass}) are likely not to hold in practice.
Hence, we explored concrete examples where \methodname and SVM-based computation differ in predictive performance. 
For example, we find in ViT-B/16 \cite{dosovitskiy2020vit}, after the encoder layer 10, an accuracy difference of 40\% with the SVM yielding an accuracy of 95\% and \methodname of 55\%. 
Overall, we observed significant accuracy differences (over 25 percentage points) in 2.8\% of the CAVs identified, favoring SVM in 1\% and \methodname in 1.8\% of cases. 
In the former case, we believe that these findings are circumstances where specifically the isotropic Gaussian assumption is violated, meaning the means of $D_c$ and $D_r$ are close together in comparison to the convex hull of both sets.
Overall, we find that \textbf{in 43.6\% of the cases \methodname achieves a higher accuracy, in 36.9\% of the cases \oldmethodname outperforms, and in 19.5\% of the cases both methods perform equally}.
These observations support the findings in \Cref{sec:experiments} in our main paper and generally favor \methodname.

\begin{figure*}[!htb]
    \begin{subfigure}[t]{0.49\textwidth}
        \centering
        \includegraphics[width=\linewidth]{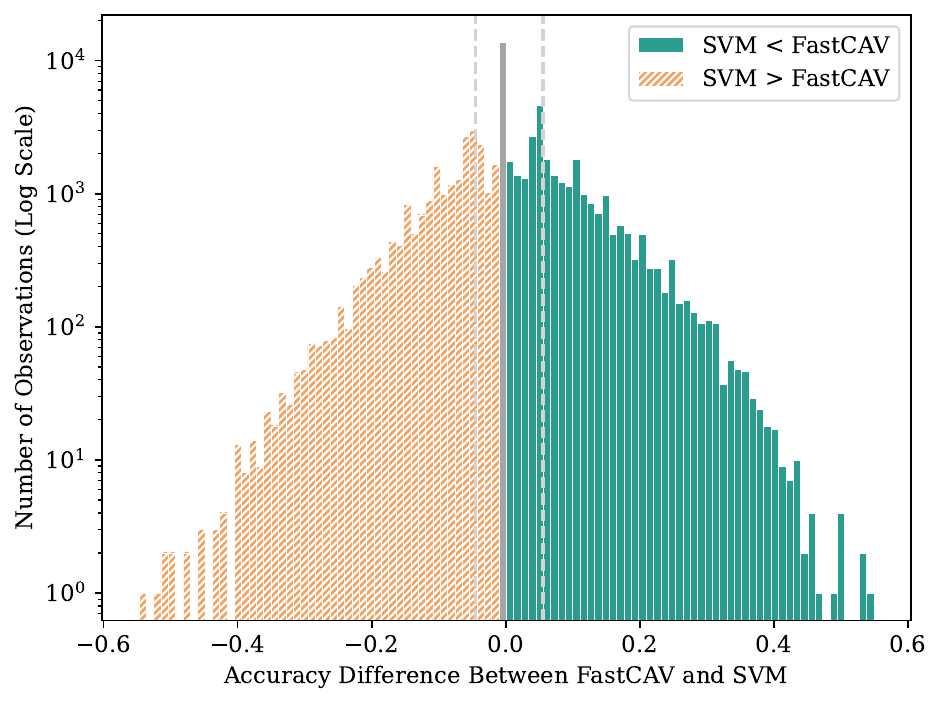}
        \label{fig:appendix:svm_samples}
    \end{subfigure}
    \hfill
    \begin{subfigure}[t]{0.49\textwidth}
        \centering
        \includegraphics[width=\linewidth]{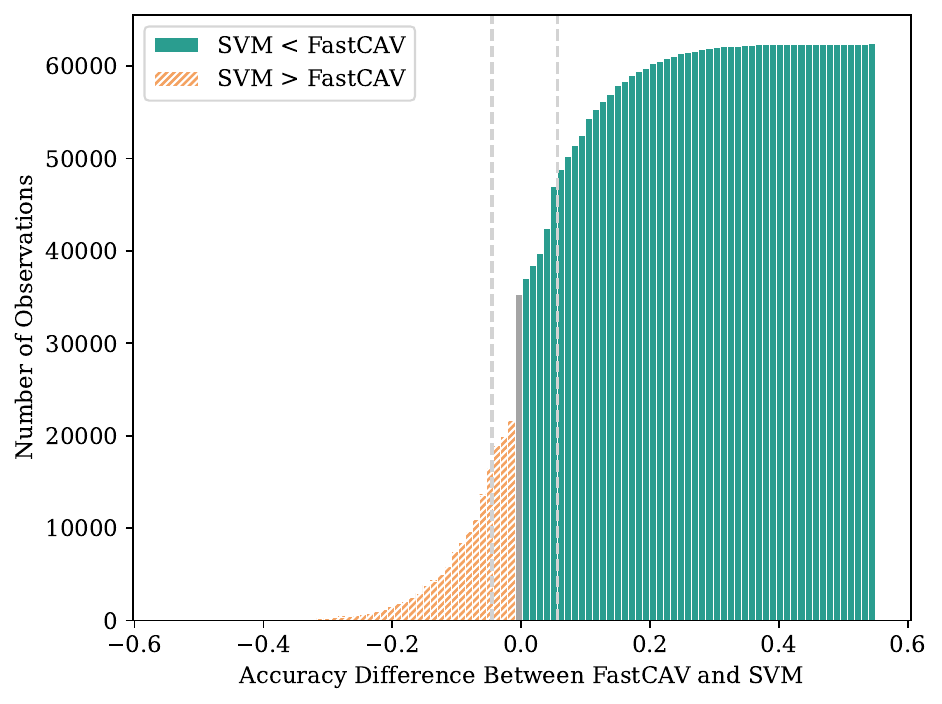}
        \label{fig:appendix:svm_dimension}
    \end{subfigure}
    \vspace{-0.2cm}
    \caption{
    Left: Histogram of accuracy differences between \methodname and \oldmethodname. 
    Note the logarithmic y-axis scale. 
    Right: Cumulative histogram (empirical CDF) of accuracy differences between \methodname and SVM. 
    Both figures show the differences for all layers of all models (see \Cref{apx:setup}). 
    Negative values indicate that \oldmethodname achieved higher accuracy, while positive values indicate that \methodname performed better. 
    Agreement between both methods (a difference of zero) is colored in grey, and the grey vertical dashed lines mark the 25th and 75th percentiles (Q1 and Q3).
    }
    \label{fig:appendix:svm_benchmark}
\end{figure*}

Next, we compare \methodname and SVM-based computation in a more specialized task aiming to extract CAVs for medical concepts \cite{johnson2019mimic-cxr,irvin2019chexpert} to show its broad applicability.

\subsubsection{CAV Computation for Medical Concepts}
\label{apx:medical}

In our main paper, we focus on general-purpose models trained on ImageNet \cite{russakovsky2015imagenet} and consider a broad set of concepts \cite{bau2017network}.
Here, we include an analysis of a more specialized second task.
Specifically, we follow a related work exploring medical concepts \cite{singla2021using} and train a DenseNet-121 \cite{huang2017densely} on a subset of MIMIC-CXR \cite{johnson2019mimic-cxr}. 

Regarding training details, we use AdamW \cite{loshchilov2017decoupled} with a learning rate of $1e-4$, a weight decay of $5e-5$, and a batch size of $64$.
We train for $100$ epochs and select the model achieving the highest multi-label AUROC (0.7735) on an independent test set for our concept-based analysis.

Next, we use the CheXpert labeler \cite{irvin2019chexpert} to extract concept mentions from the included free-text radiology reports included with MIMIC-CXR \cite{johnson2019mimic-cxr}.
As a set of possible concepts concerning chest x-ray images, we again follow \cite{singla2021using}.

With these extracted concepts, we compute CAVs for the four blocks of the DenseNet-121 using our FastCAV, SVM-based computation \cite{ghosh2023distilling}, and sparse logistic regression \cite{singla2021using}.

In our main paper (in \Cref{tab:compareFastCAVMedicalImagesSummary}), we report the average computation time, CAV accuracy, and intra-method similarity.
Here we provide a more detailed evaluation of the specific concepts after the dense-block 3, following \cite{singla2021using}.

\begin{table*}[!tb] %
    \small
    \centering
    \caption{Comparing our approach \methodname with \oldmethodname and sparse logistic regression for concepts concerning chest x-ray images.
    \textbf{Bold values} indicate better results.
    Leading zeros were removed to fit the space constraints.
    }
    \label{tab:compareFastCAVMedicalImagesConcepts}
    \begin{tabular}{lrrrrrrrrr}
        \toprule
           & \multicolumn{3}{c}{Comp. Time [$s$] $\downarrow$} & \multicolumn{3}{c}{Accuracy $\uparrow$} & \multicolumn{3}{c}{Similarity $\uparrow$} \\
          \cmidrule(lr){2-4} \cmidrule(lr){5-7} \cmidrule(lr){8-10}
         Concept & FastCAV & S-LR & SVM & FastCAV & S-LR & SVM & FastCAV & S-LR & SVM \\
        \midrule
        Air Bronchogram & \textbf{.007{\scriptsize$\pm$.000}} & 5.93{\scriptsize$\pm$0.28}  & .424{\scriptsize$\pm$.05}  & .71{\scriptsize$\pm$.09} & \textbf{.75{\scriptsize$\pm$.07}}  & .73{\scriptsize$\pm$.14}  & \textbf{.87{\scriptsize$\pm$.02}} & .49{\scriptsize$\pm$.03}  & .43{\scriptsize$\pm$.04}  \\
        Ards & \textbf{.007{\scriptsize$\pm$.000}} & 5.97{\scriptsize$\pm$0.54}  & .391{\scriptsize$\pm$.05}  & \textbf{.82{\scriptsize$\pm$.11}} & .79{\scriptsize$\pm$.07}  & .76{\scriptsize$\pm$.14}  & \textbf{.89{\scriptsize$\pm$.01}} & .54{\scriptsize$\pm$.04}  & .48{\scriptsize$\pm$.04}  \\
        Blunt & \textbf{.006{\scriptsize$\pm$.001}} & 6.71{\scriptsize$\pm$0.79}  & .441{\scriptsize$\pm$.02}  & \textbf{.72{\scriptsize$\pm$.13}} & .59{\scriptsize$\pm$.07}  & .68{\scriptsize$\pm$.15}  & \textbf{.85{\scriptsize$\pm$.03}} & .43{\scriptsize$\pm$.05}  & .40{\scriptsize$\pm$.05}  \\
        Cardiac Silhouette & \textbf{.006{\scriptsize$\pm$.000}} & 5.87{\scriptsize$\pm$0.44}  & .476{\scriptsize$\pm$.07}  & .70{\scriptsize$\pm$.07} & .69{\scriptsize$\pm$.09}  & \textbf{.71{\scriptsize$\pm$.06}}  & \textbf{.82{\scriptsize$\pm$.03}} & .49{\scriptsize$\pm$.04}  & .46{\scriptsize$\pm$.03}  \\
        Cephalization & \textbf{.007{\scriptsize$\pm$.000}} & 6.36{\scriptsize$\pm$0.54}  & .405{\scriptsize$\pm$.04}  & .72{\scriptsize$\pm$.10} & \textbf{.80{\scriptsize$\pm$.13}}  & .80{\scriptsize$\pm$.06}  & \textbf{.85{\scriptsize$\pm$.02}} & .57{\scriptsize$\pm$.02}  & .49{\scriptsize$\pm$.03}  \\
        Congestion & \textbf{.006{\scriptsize$\pm$.001}} & 6.14{\scriptsize$\pm$0.45}  & .424{\scriptsize$\pm$.08}  & .78{\scriptsize$\pm$.06} & .78{\scriptsize$\pm$.07}  & \textbf{.80{\scriptsize$\pm$.09}}  & \textbf{.86{\scriptsize$\pm$.02}} & .48{\scriptsize$\pm$.03}  & .49{\scriptsize$\pm$.04}  \\
        Heart Size & \textbf{.007{\scriptsize$\pm$.000}} & 6.18{\scriptsize$\pm$0.64}  & .435{\scriptsize$\pm$.07}  & .72{\scriptsize$\pm$.10} & .73{\scriptsize$\pm$.10}  & \textbf{.76{\scriptsize$\pm$.06}}  & \textbf{.81{\scriptsize$\pm$.03}} & .47{\scriptsize$\pm$.05}  & .45{\scriptsize$\pm$.05}  \\
        Hilar Contour & \textbf{.007{\scriptsize$\pm$.000}} & 6.71{\scriptsize$\pm$1.20}  & .442{\scriptsize$\pm$.02}  & \textbf{.71{\scriptsize$\pm$.10}} & .69{\scriptsize$\pm$.08}  & .67{\scriptsize$\pm$.08}  & \textbf{.78{\scriptsize$\pm$.02}} & .42{\scriptsize$\pm$.04}  & .42{\scriptsize$\pm$.03}  \\
        Hilar Engorgement & \textbf{.006{\scriptsize$\pm$.001}} & 6.62{\scriptsize$\pm$0.81}  & .370{\scriptsize$\pm$.04}  & .75{\scriptsize$\pm$.10} & \textbf{.82{\scriptsize$\pm$.07}}  & .72{\scriptsize$\pm$.10}  & \textbf{.85{\scriptsize$\pm$.02}} & .54{\scriptsize$\pm$.05}  & .48{\scriptsize$\pm$.04}  \\
        Hilar Opacity & \textbf{.007{\scriptsize$\pm$.000}} & 7.01{\scriptsize$\pm$0.62}  & .451{\scriptsize$\pm$.05}  & .68{\scriptsize$\pm$.13} & \textbf{.75{\scriptsize$\pm$.06}}  & .63{\scriptsize$\pm$.05}  & \textbf{.79{\scriptsize$\pm$.03}} & .47{\scriptsize$\pm$.05}  & .43{\scriptsize$\pm$.05}  \\
        Interst. Edema & \textbf{.006{\scriptsize$\pm$.000}} & 6.23{\scriptsize$\pm$0.81}  & .414{\scriptsize$\pm$.07}  & .82{\scriptsize$\pm$.07} & \textbf{.83{\scriptsize$\pm$.06}}  & .80{\scriptsize$\pm$.05}  & \textbf{.89{\scriptsize$\pm$.01}} & .52{\scriptsize$\pm$.04}  & .49{\scriptsize$\pm$.04}  \\
        Interst. Markings & \textbf{.006{\scriptsize$\pm$.001}} & 6.20{\scriptsize$\pm$0.79}  & .463{\scriptsize$\pm$.05}  & .65{\scriptsize$\pm$.09} & .70{\scriptsize$\pm$.10}  & \textbf{.71{\scriptsize$\pm$.10}}  & \textbf{.83{\scriptsize$\pm$.02}} & .51{\scriptsize$\pm$.03}  & .44{\scriptsize$\pm$.04}  \\
        Interst. Prominence & \textbf{.005{\scriptsize$\pm$.000}} & 6.72{\scriptsize$\pm$0.64}  & .427{\scriptsize$\pm$.05}  & .73{\scriptsize$\pm$.16} & .78{\scriptsize$\pm$.16}  & \textbf{.78{\scriptsize$\pm$.05}}  & \textbf{.86{\scriptsize$\pm$.02}} & .46{\scriptsize$\pm$.03}  & .42{\scriptsize$\pm$.02}  \\
        Peribronch. Cuffing & \textbf{.007{\scriptsize$\pm$.000}} & 6.37{\scriptsize$\pm$0.20}  & .478{\scriptsize$\pm$.07}  & \textbf{.72{\scriptsize$\pm$.05}} & .65{\scriptsize$\pm$.06}  & .69{\scriptsize$\pm$.13}  & \textbf{.72{\scriptsize$\pm$.04}} & .51{\scriptsize$\pm$.02}  & .48{\scriptsize$\pm$.02}  \\
        Pleural Fluid & \textbf{.006{\scriptsize$\pm$.001}} & 6.54{\scriptsize$\pm$1.33}  & .434{\scriptsize$\pm$.03}  & \textbf{.76{\scriptsize$\pm$.13}} & .72{\scriptsize$\pm$.09}  & .72{\scriptsize$\pm$.12}  & \textbf{.82{\scriptsize$\pm$.02}} & .45{\scriptsize$\pm$.04}  & .43{\scriptsize$\pm$.05}  \\
        Vascular Marking & \textbf{.006{\scriptsize$\pm$.001}} & 5.74{\scriptsize$\pm$0.87}  & .439{\scriptsize$\pm$.05}  & \textbf{.81{\scriptsize$\pm$.07}} & .79{\scriptsize$\pm$.09}  & .74{\scriptsize$\pm$.13}  & \textbf{.85{\scriptsize$\pm$.02}} & .49{\scriptsize$\pm$.04}  & .46{\scriptsize$\pm$.03}  \\
        \bottomrule
    \end{tabular}
\end{table*}

\Cref{tab:compareFastCAVMedicalImagesConcepts} summarizes the computation time, CAV accuracies, and intra-class similarities for the set of concepts.
Overall, we find that the DenseNet-121 \cite{huang2017densely} learned the specialized concepts, and we find high accuracies for all three computation methods.
However, the specific accuracies per concept vary slightly.
Further, we see differences in the average similarity for the three methods.
\methodname again performs best with respect to speed and similarity.

Additionally, we investigate CAV computation with respect to images containing negative results for the concepts as found by the CheXpert labeler \cite{irvin2019chexpert}. 
In contrast to our expectation, we observe lower CAV accuracies in comparison to random images. 
However, this also holds for SVM and sparse-logistic regression-based computation, indicating a data sampling problem in the negative example regime.

Lastly, following the similar results to established methods, we wish to highlight the applicability of \methodname for specialized downstream tasks, e.g., \cite{singla2021using,ghosh2023distilling}.
We consider this a valuable direction for future work.

\subsection{TCAV --- Additional Results}
\label{apx:tcav}

In our main paper, we include the results for TCAV \cite{kim2018interpretabilityfeatureattributionquantitative} for the class \emph{ladybug} using both our \methodname as well as SVM-based CAV computation.
To further strengthen our analysis, we follow \cite{kim2018interpretabilityfeatureattributionquantitative} and evaluate additional qualitative examples.
Here, we follow the setup detailed in \Cref{subsec:xai_methods} and use a GoogleNet \cite{szegedy2015going} trained on ImageNet \cite{russakovsky2015imagenet}.
We visualize the TCAV scores for the same layers selected in \cite{kim2018interpretabilityfeatureattributionquantitative}.

\begin{figure*}[!htb]
    \centering
    \includegraphics[width=\linewidth]{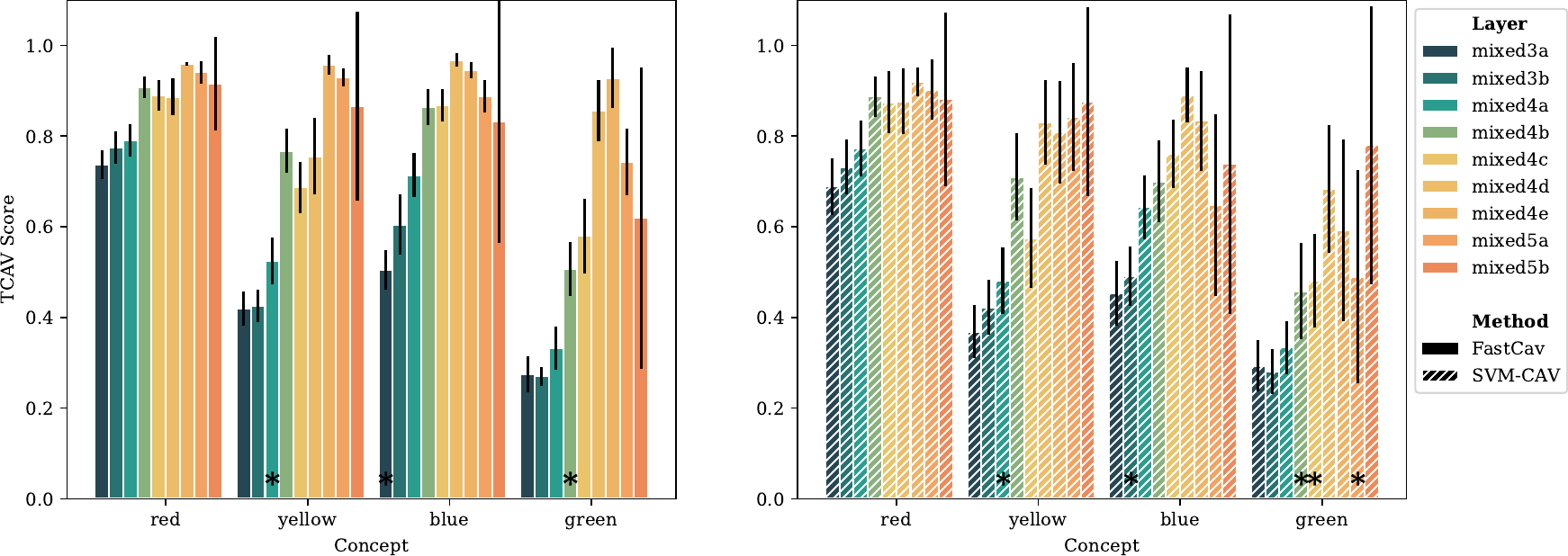}
    \caption{Example of the concepts \concept{red}, \concept{yellow}, \concept{blue} and \concept{green} for class \emph{ladybug} using TCAV.
    We follow \cite{kim2018interpretabilityfeatureattributionquantitative} and mark CAVs that are not statistically significant with ``*''.
    }
    \label{fig:tcav:ladybug:method}
\end{figure*}

\begin{figure*}[!htb]
    \centering
    \includegraphics[width=\linewidth]{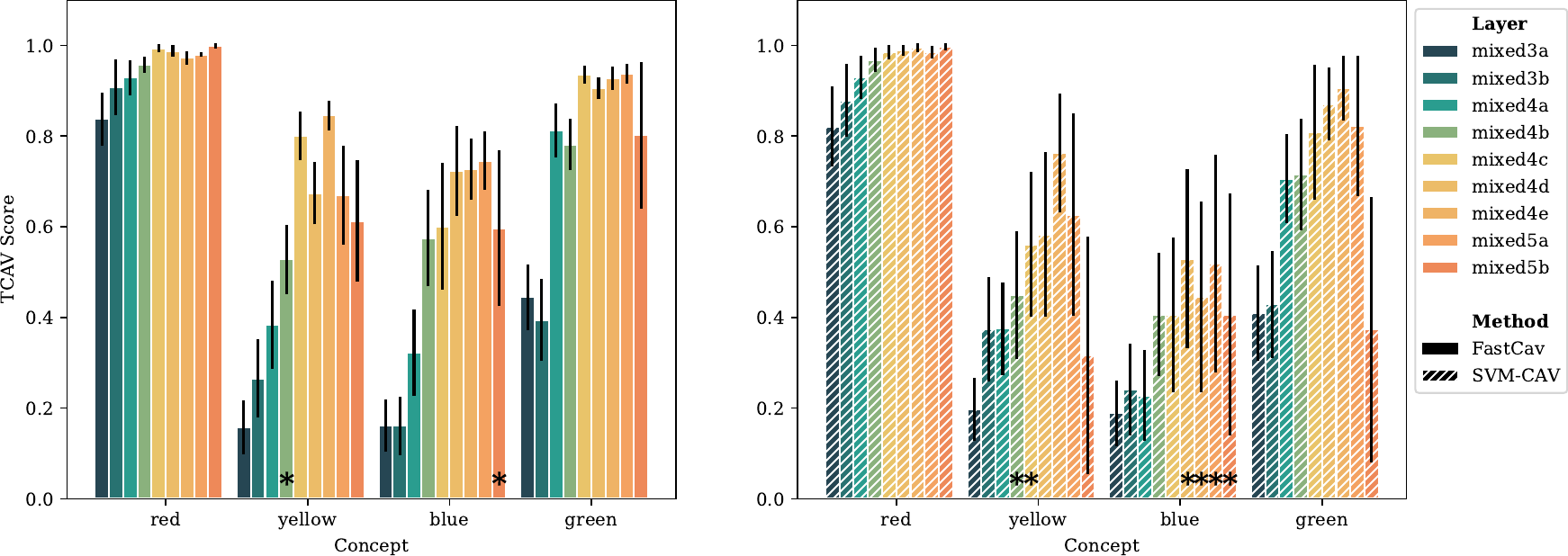}
    \caption{Example of the concepts \concept{red}, \concept{yellow}, \concept{blue} and \concept{green} for class \emph{fire-engine} using TCAV.
    We follow \cite{kim2018interpretabilityfeatureattributionquantitative} and mark CAVs that are not statistically significant with ``*''.
    }
    \label{fig:tcav:fire-engine:method}
\end{figure*}

\begin{figure*}[!htb]
    \centering
    \includegraphics[width=\linewidth]{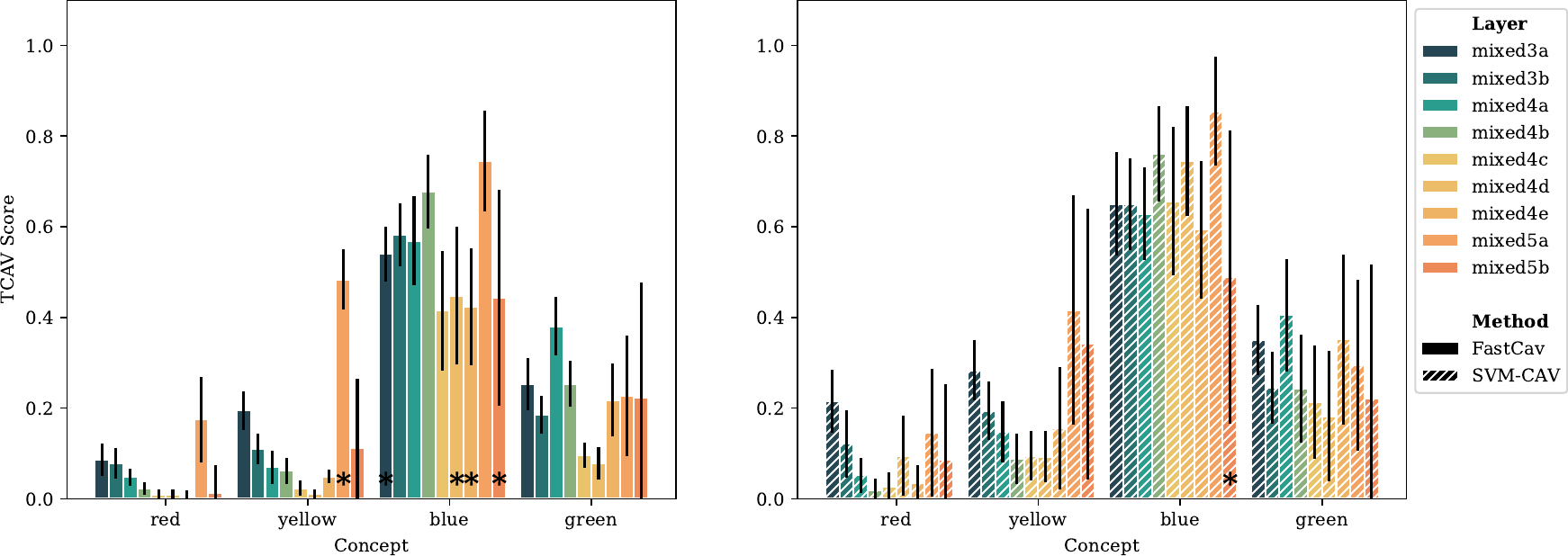}
    \caption{Example of the concepts \concept{red}, \concept{yellow}, \concept{blue} and \concept{green} for class \emph{police-van} using TCAV.
    We follow \cite{kim2018interpretabilityfeatureattributionquantitative} and mark CAVs that are not statistically significant with ``*''.
    }
    \label{fig:tcav:police:method}
\end{figure*}

\begin{figure*}[!tb]
    \centering
    \includegraphics[width=\linewidth]{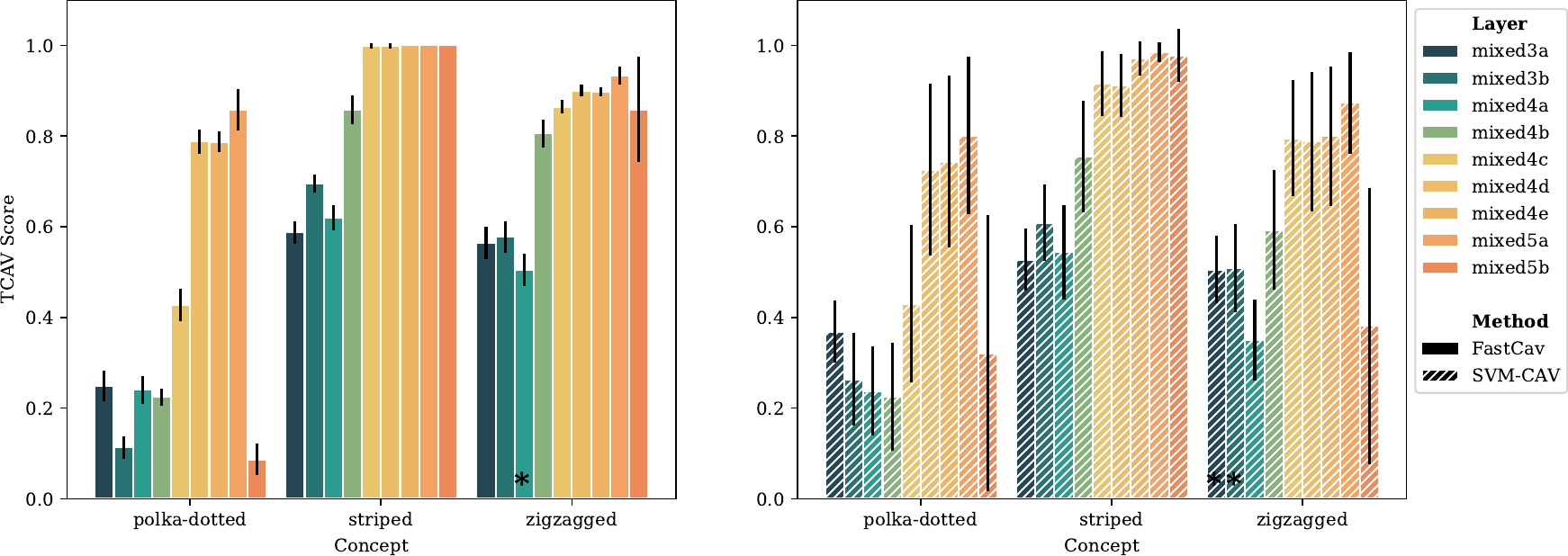}
    \caption{Example of the concepts \concept{polka-dotted}, \concept{striped} and \concept{zigzagged} for class \emph{zebra} using TCAV.
    We follow \cite{kim2018interpretabilityfeatureattributionquantitative} and mark CAVs that are not statistically significant with ``*''.
    }
    \label{fig:tcav:zebra:method}
\end{figure*}

\Cref{fig:tcav:ladybug:method}, \Cref{fig:tcav:fire-engine:method}, \Cref{fig:tcav:police:method}, and \Cref{fig:tcav:zebra:method} display classes \emph{ladybug}, \emph{fire-engine}, \emph{police-van}, and \emph{zebra}, respectively.
In all cases, we find only small differences between our CAV computation and the established SVM-based approach.
Similar to our discussion in the main paper, we find lower variances using \methodname.
Nevertheless, the overarching trends are consistent between both methods.
Further, the resulting insights into the GoogleNet model align well with \cite{kim2018interpretabilityfeatureattributionquantitative}.
For example, for both methods, we find the concept \concept{red} to be most important for the classes \emph{ladybug} and \emph{fire-engine}.
Similarly, we observe \concept{blue} influencing class \emph{police-van} and \concept{striped} influencing class \emph{zebra}.
This again demonstrates that both \methodname and \oldmethodname can be used to generate valid concept-based explanations, supporting the claims made in our main paper.


\FloatBarrier

\subsection{ACE --- Additional Results}
\label{apx:ace}

\begin{table}[!tb]
    \centering
    \caption{
    The TCAV scores \cite{kim2018interpretabilityfeatureattributionquantitative} for the four most salient concepts discovered in an ImageNet \cite{russakovsky2015imagenet} trained Inception-v3 \cite{szegedy2016rethinking} model using ACE \cite{ghorbani2019towards}.
    We list scores for the classes \emph{Lionfish}, \emph{Zebra}, and \emph{Police Van}. 
    }
    \label{tab:ace_tcav_scores}
    \begin{tabular}{lcc:cc:cc}
        \toprule
        & \multicolumn{2}{c}{\emph{Lionfish}} & \multicolumn{2}{c}{\emph{Zebra}} & \multicolumn{2}{c}{\emph{Police Van}}\\
        \cmidrule(lr){2-3} \cmidrule(lr){4-5} \cmidrule(lr){6-7}
        Concept  & \methodname & \oldmethodname & \methodname & \oldmethodname & \methodname & \oldmethodname\\
        \midrule
        1  & 0.78 & 0.73 & 0.68 & 0.66 & 0.88 & 0.78\\
        2  & 0.75 & 0.73 & 0.67 & 0.61 & 0.86 & 0.77\\
        3  & 0.69 & 0.67 & 0.60 & 0.58 & 0.83 & 0.76\\
        4  & 0.68 & 0.67 & 0.59 & 0.56 & 0.80 & 0.70\\
        \bottomrule
    \end{tabular}
\end{table}

In our main paper, we visualize and compare the most salient concepts for the class \emph{Lionfish} found by ACE with \methodname and \oldmethodname.
Here, we add to this and visualize the other classes selected in \cite{ghorbani2019towards}.
For class \emph{Lionfish}, ACE using \methodname finds 22 relevant concepts, and with \oldmethodname, we identify 24 overall.
For \emph{Zebra}, we find 19 concepts using our approach and 18 with SVM computation.
Lastly, for class \emph{Police Van}, we identify 24 and 22, respectively.
In \Cref{tab:ace_tcav_scores}, we list the TCAV scores of the four most salient concepts for each class, respectively.

\begin{figure*}[!htb]
    \begin{subfigure}[t]{0.49\textwidth}
        \centering
        \includegraphics[width=\linewidth]{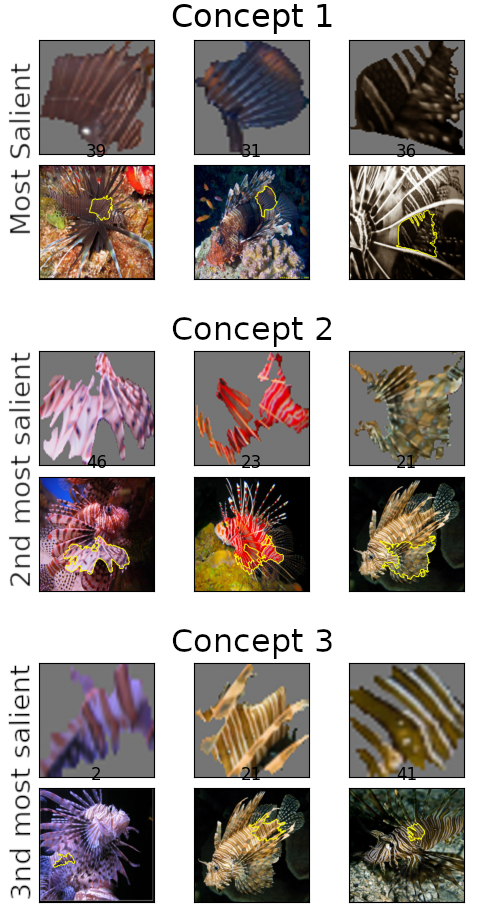}
        \caption{
        \methodname based ACE \cite{ghorbani2019towards}.
        }
        \label{fig:ace_lionfish_mean}
    \end{subfigure}
    \hfill
    \begin{subfigure}[t]{0.49\textwidth}
        \centering
        \includegraphics[width=\linewidth]{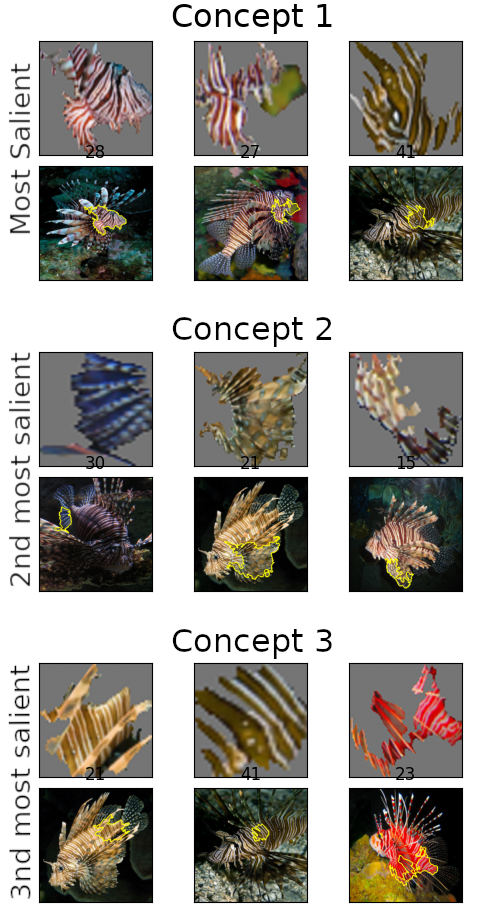}
        \caption{\oldmethodname based ACE \cite{ghorbani2019towards}.}
        \label{fig:ace_lionfish_cav}
    \end{subfigure}
    \vspace{-0.2cm}
    \caption{
    Comparison of the most salient concepts discovered by ACE \cite{ghorbani2019towards} using either our \methodname or the established \oldmethodname.
    Here, we use class \emph{lionfish} and display the three most salient concepts.
    We find the discovered patches between both approaches similar and congruent with the original observation in \cite{ghorbani2019towards}.
    }
    \label{fig:ace_lionfish}
\end{figure*}

\begin{figure*}[!htb]
    \begin{subfigure}[t]{0.49\textwidth}
        \centering
        \includegraphics[width=\linewidth]{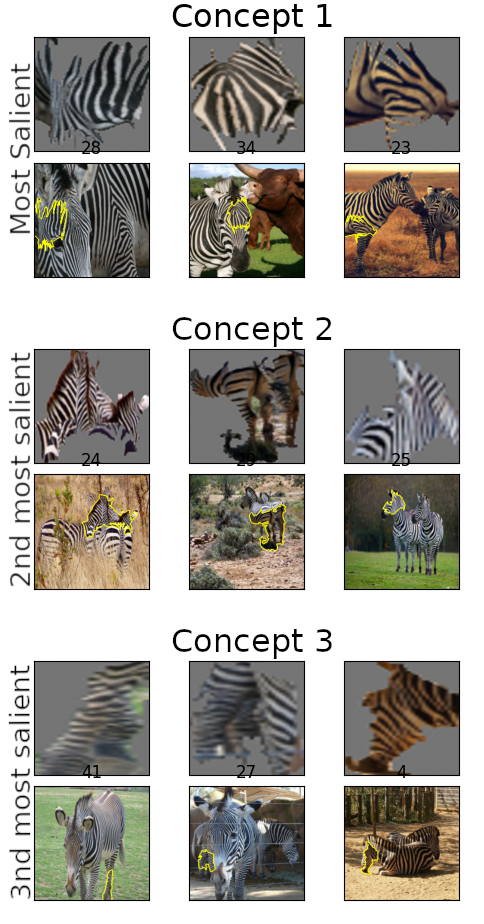}
        \caption{
        \methodname based ACE \cite{ghorbani2019towards}.
        }
        \label{fig:ace_zebra_mean}
    \end{subfigure}
    \hfill
    \begin{subfigure}[t]{0.49\textwidth}
        \centering
        \includegraphics[width=\linewidth]{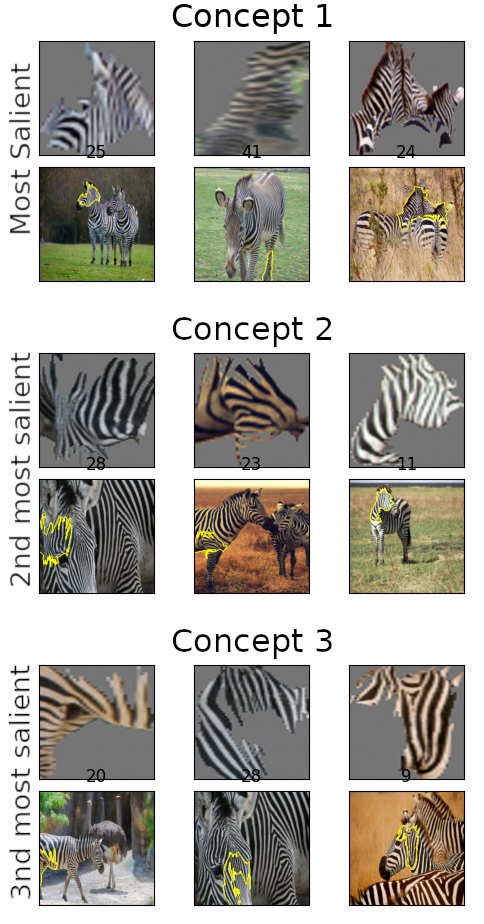}
        \caption{\oldmethodname based ACE \cite{ghorbani2019towards}.}
        \label{fig:ace_zebra_cav}
    \end{subfigure}
    \vspace{-0.2cm}
    \caption{
    Comparison of the most salient concepts discovered by ACE \cite{ghorbani2019towards} using either our \methodname or the established \oldmethodname.
    Here, we use class \emph{zebra} and display the three most salient concepts.
    We find the discovered patches between both approaches similar and congruent with the original observation in \cite{ghorbani2019towards}.
    }
    \label{fig:ace_zebra}
\end{figure*}

\begin{figure*}[!htb]
    \begin{subfigure}[t]{0.49\textwidth}
        \centering
        \includegraphics[width=\linewidth]{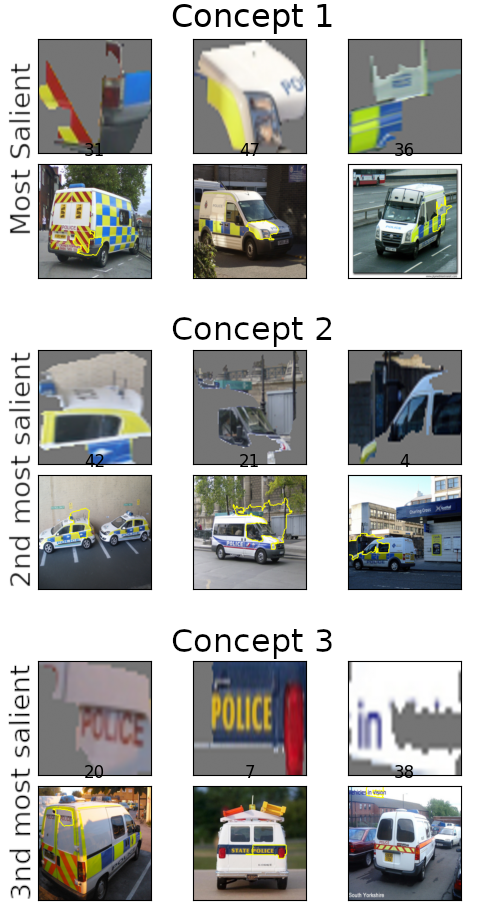}
        \caption{
        \methodname based ACE \cite{ghorbani2019towards}.
        }
        \label{fig:ace_police_mean}
    \end{subfigure}\hfill
    \begin{subfigure}[t]{0.49\textwidth}
        \centering
        \includegraphics[width=\linewidth]{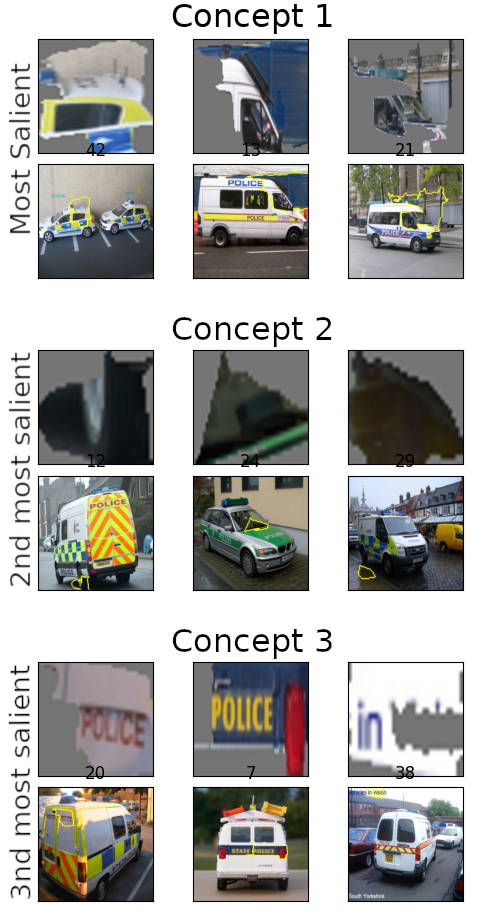}
        \caption{
        \oldmethodname based ACE \cite{ghorbani2019towards}.
        }
        \label{fig:ace_police_cav}
    \end{subfigure}
    \vspace{-0.2cm}
    \caption{
    Comparison of the most salient concepts discovered by ACE \cite{ghorbani2019towards} using either our \methodname or the established \oldmethodname.
    Here, we use class \emph{police} and display the three most salient concepts.
    We find the discovered patches between both approaches similar and congruent with the original observation in \cite{ghorbani2019towards}.
    }
    \label{fig:ace_police}
\end{figure*}

In \Cref{fig:ace_lionfish}, \Cref{fig:ace_zebra}, and \Cref{fig:ace_police}, we display the three most salient concepts for both methodologies.
In all cases, we observe similar results for both \methodname and \oldmethodname, supporting our findings in the main paper.

\FloatBarrier
\subsection{Tracking CAVs During Training --- Additional Visualizations}
\label{appendix:tracking_cavs}

\paragraph{Setup Details:} 
We train a ResNet50 \cite{he2016deep} model on the ImageNet 2012 dataset \cite{russakovsky2015imagenet} using a batch size of 256 for 90 epochs following \cite{paszke2019pytorch}.
Each image is resized to 256 pixels, followed by a random crop of 224 by 224, a random horizontal flip, and normalization with mean values of 0.485, 0.456, and 0.406 and standard deviations of 0.229, 0.224, and 0.225. 
The optimizer is Stochastic Gradient Descent with an initial learning rate of 0.1, momentum of 0.9, weight decay of 1e-4, and a StepLR scheduler with a step size of 30 epochs and a gamma of 0.1.

\paragraph{Additional Visualizations:}

\begin{figure}[ht]
    \centering
    \begin{subfigure}{0.49\linewidth}
        \includegraphics[width=\textwidth]{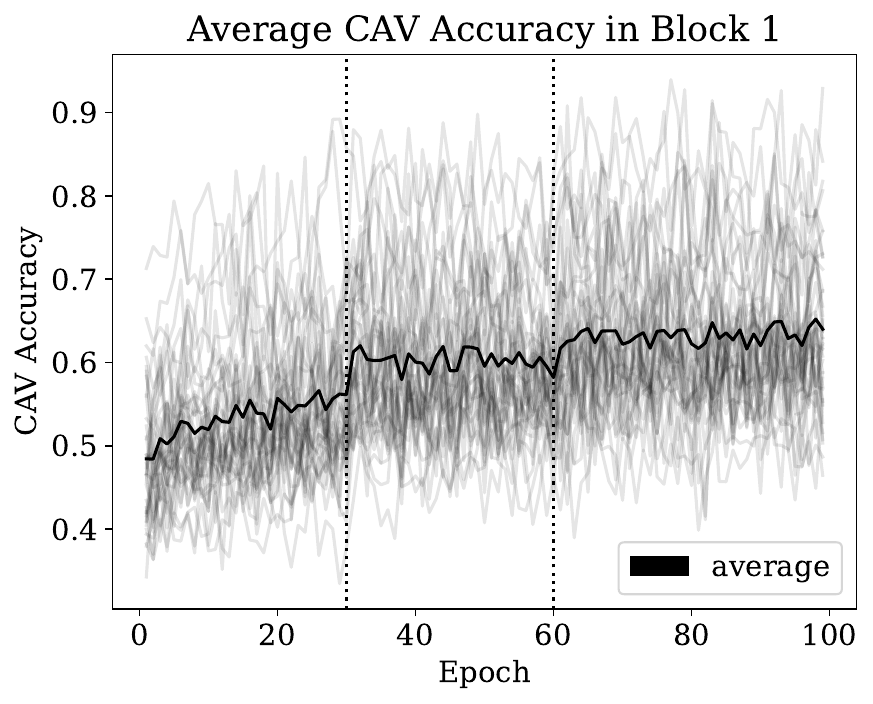} %
        
        \caption{Final layer ResNet50 \textbf{Block 1}.}
        \label{fig:avg_block1}
    \end{subfigure}
    \begin{subfigure}{0.49\linewidth}
        \includegraphics[width=\textwidth]{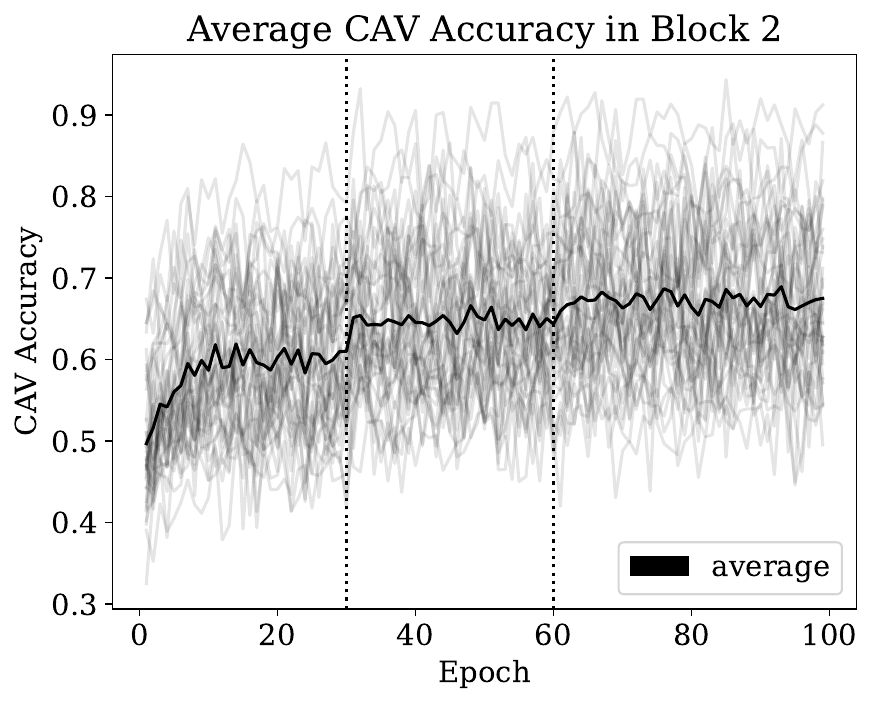} %
        
        \caption{Final layer ResNet50 \textbf{Block 2}}
        \label{fig:avg_block2}
    \end{subfigure}
    \\
    
    \begin{subfigure}{0.49\linewidth}
        \includegraphics[width=\textwidth]{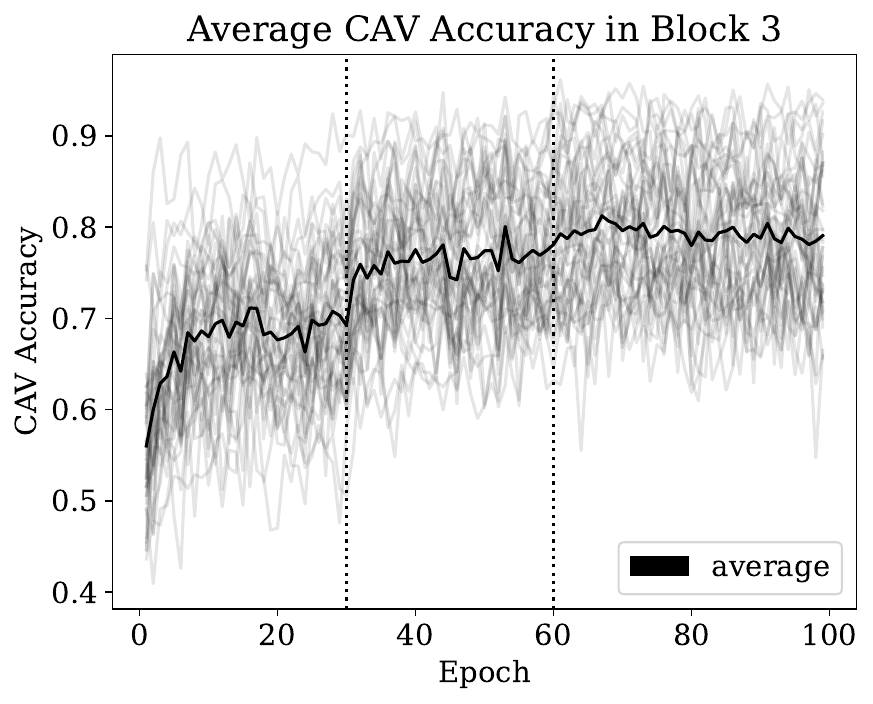} %
        
        \caption{Final layer ResNet50 \textbf{Block 3}.}
        \label{fig:avg_block3}
    \end{subfigure}
    \begin{subfigure}{0.49\linewidth}
        \includegraphics[width=\textwidth]{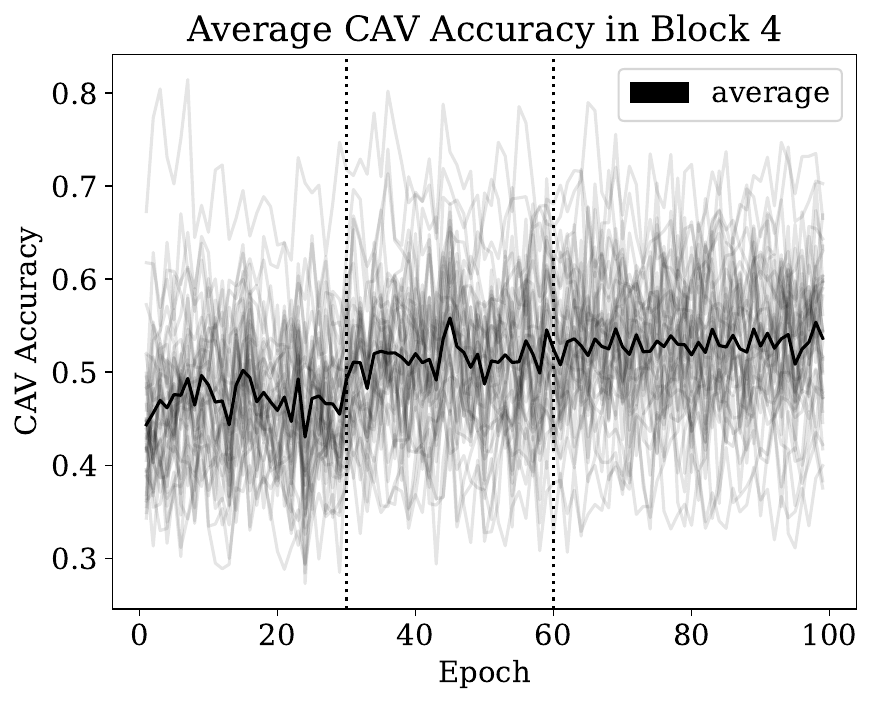} %
        
        \caption{Final layer ResNet50 \textbf{Block 4}}
        \label{fig:avg_block4}
    \end{subfigure}
    \caption{
    Additional visualizations pertaining to the lefthand side in \Cref{fig:cavs_during_training}.
    Specifically, we visualize the average accuracies achieved by CAVs after the final layers in each of the four ResNet blocks (\Cref{fig:avg_block1} - \Cref{fig:avg_block4}) for the concepts in \cite{bau2017network}.
    Here, we display the changes for all concepts to showcase the respective variances during training.
    Vertical dotted lines again correspond to epochs where the learning rate was divided by ten.
    }
    \label{fig:apx:avg_resnet}
\end{figure}

In our main paper (\Cref{fig:cavs_during_training}), we visualize the average CAV accuracies during the training of a ResNet50 \cite{he2016deep}.
We calculate this average over the concepts included in \cite{bau2017network}.
In \Cref{fig:apx:avg_resnet}, we visualize the same averages but additionally include lines for all individual concepts to highlight the variances.

\begin{figure}[ht]
    \centering
    \begin{subfigure}{0.49\linewidth}
        \includegraphics[width=\textwidth]{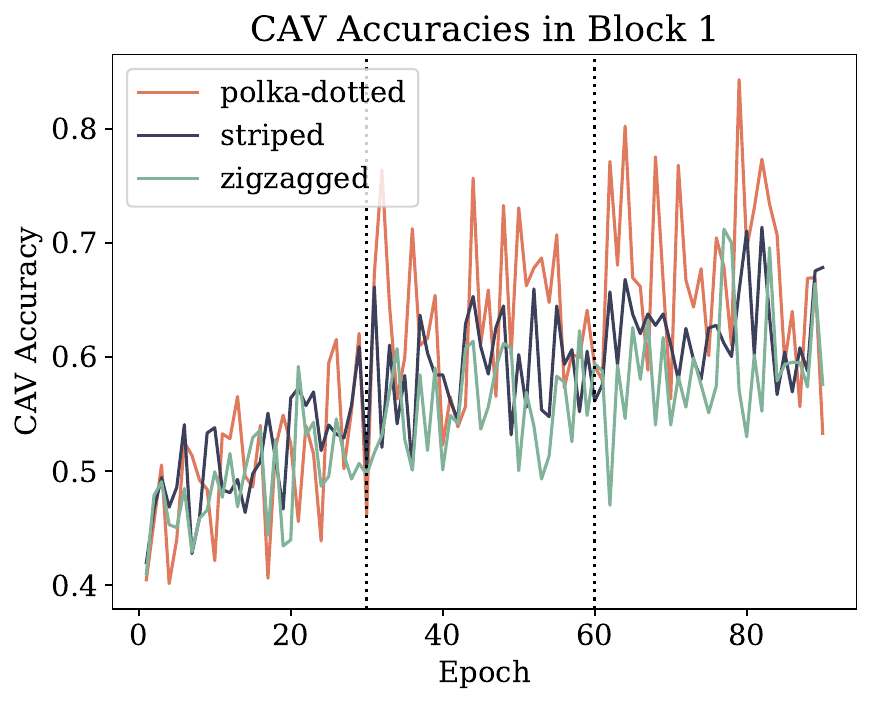} %
        
        \caption{Final layer ResNet50 \textbf{Block 1}.}
        \label{fig:concept_block1}
    \end{subfigure}
    \begin{subfigure}{0.49\linewidth}
        \includegraphics[width=\textwidth]{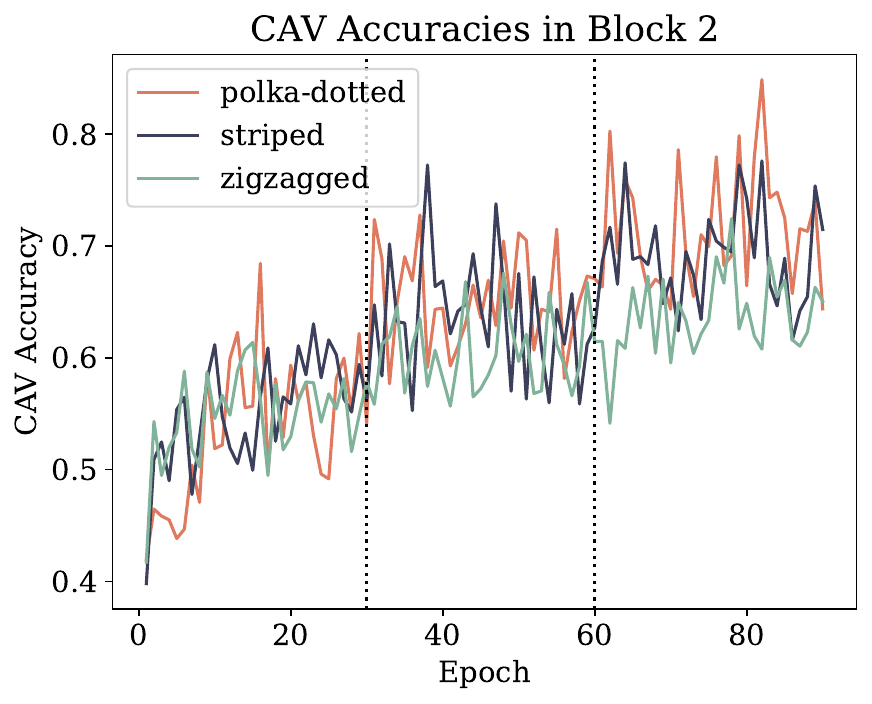} %
        
        \caption{Final layer ResNet50 \textbf{Block 2}}
        \label{fig:concept_block2}
    \end{subfigure}
    \\
    
    \begin{subfigure}{0.49\linewidth}
        \includegraphics[width=\textwidth]{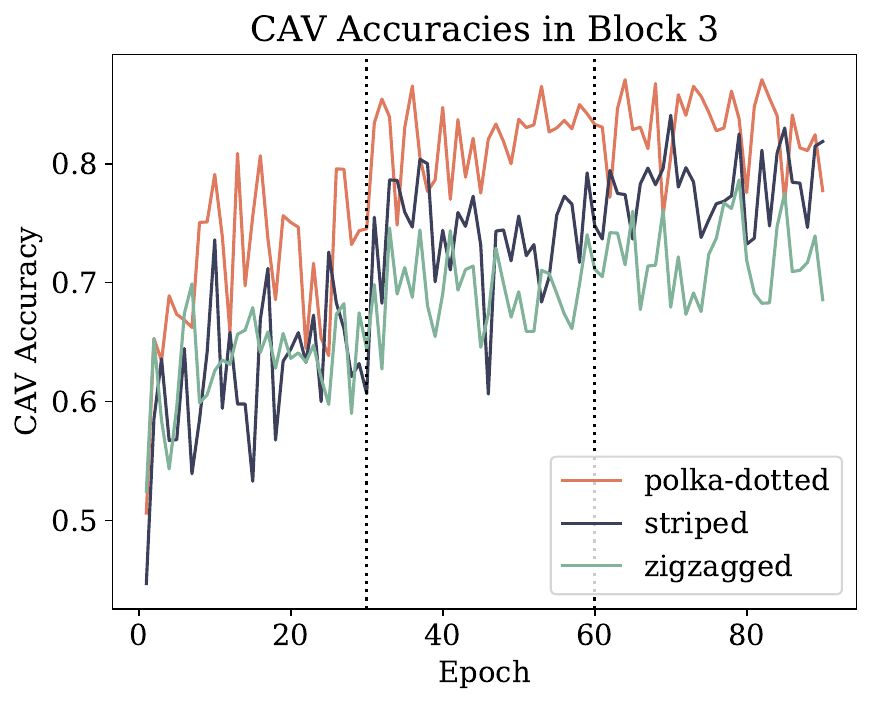} %
        
        \caption{Final layer ResNet50 \textbf{Block 3}}
        \label{fig:concept_block3}
    \end{subfigure}
    \begin{subfigure}{0.49\linewidth}
        \includegraphics[width=\textwidth]{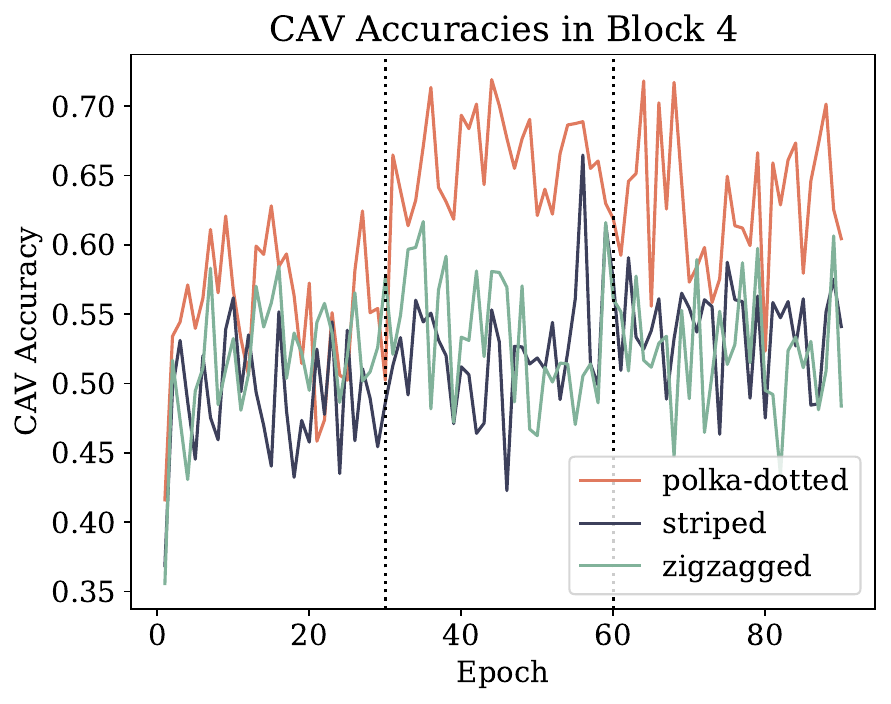} %
        
        \caption{Final layer ResNet50 \textbf{Block 4}.}
        \label{fig:concept_block4}
    \end{subfigure}
    \caption{
    Additional visualizations for the middle plot in \Cref{fig:cavs_during_training}.
    Specifically, we qualitatively visualize the accuracies of selected CAVs after the final layers in each of the four ResNet blocks (\Cref{fig:concept_block1} - \Cref{fig:concept_block4}) for the concepts in \cite{bau2017network}.
    \Cref{fig:concept_block3} is equivalent to the selected version in \Cref{fig:cavs_during_training}.
    Vertical dotted lines again correspond to epochs where the learning rate was divided by ten.
    }
    \label{fig:apx:concept_resnet}
\end{figure}

Similarly, following our qualitative selection of specific concepts for the final layer of block 3 in \Cref{fig:cavs_during_training} (middle plot), we include visualizations for the other blocks in \Cref{fig:apx:concept_resnet}.
We confirm that the observations made in the main paper similarly hold for the other layers.
In addition, we find the lowest CAV accuracies in \Cref{fig:concept_block4}, which is consistent with the results for the development of the average concept.

\paragraph{Additional Explorative Analysis}

To expand our discussion regarding the training analysis (see \Cref{subsec:training-analysis} and above), we provide a more comprehensive exploration of the potential of \methodname.

\begin{figure}[ht]
    \centering
    \includegraphics[]{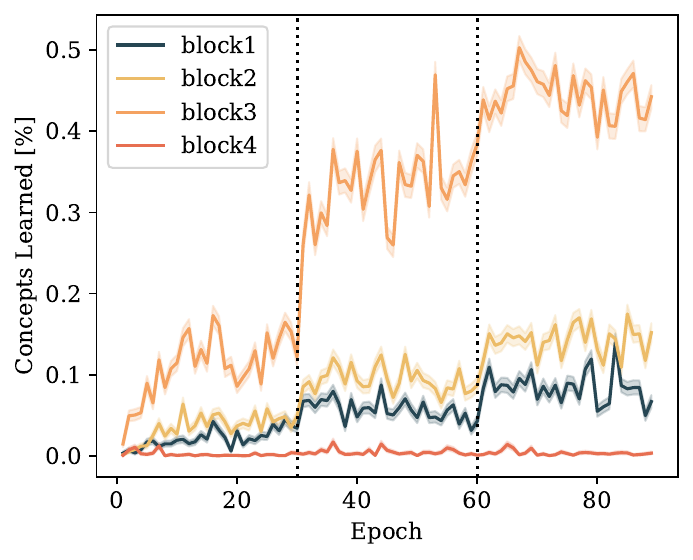}
    \caption{Percentage of learned concepts by CAVs and block during training of a ResNet50 \cite{he2016deep} on ImageNet \cite{russakovsky2015imagenet}.
    }
    \label{apx:learned_concepts}
\end{figure}

First, in our main paper, we include results ranking concepts by their area under the curve (AUC) during training and across layers. 
This reveals that the learnability/difficulty of concepts can vary wildly across layers, with some concepts that are quickly learned in early layers becoming more difficult to learn in later layers. 

Here, we go one step further and evaluate the ratio of learned concepts per layer during the training.
In \cref{apx:learned_concepts}, we summarize the corresponding results. 
We find that during the training, the ratio of concepts learned increases, indicating that the model learns task-relevant concepts.
These findings are congruent with previous analysis of neural network training dynamics, e.g., \cite{shwartz2017opening,penzel2022investigating}. 
Further, we observe the highest ratios during the complete training process for the architecture block 3.
Given that our probing dataset focuses on the textures, this observation aligns with findings made in \cite{bau2017network}.

Overall, we hope that by providing a faster and more efficient concept extraction procedure with \methodname, we can empower researchers to explore more complex scenarios and larger concept sets, potentially uncovering novel phenomena that were previously inaccessible due to computational constraints.

\end{document}